# Time Series Diffusion Method: A Denoising Diffusion Probabilistic Model for Vibration Signal Generation

Haiming Yi[a], Lei Hou[a*], Yuhong Jin[a], Nasser A. Saeed[b,c,d], Ali Kandil[b,e], Hao Duan[f]

*a) School of Astronautics, Harbin Institute of Technology, Harbin 150001, China;*
*b) Department of Physics and Engineering Mathematics, Faculty of Electronic Engineering, Menoufia University, 32952 Menouf, Egypt;*
*c) Department of Automation, Biomechanics, and Mechatronics, Faculty*
*d) Department of Mechanical Engineering, Lodz University of Technology, 90924 Lodz, Poland;*
*e) Mathematics Department, Faculty of Science, Galala University, 43511 Galala City, Egypt;*
*f) Shenyang Blower Works Group Corporation, Shenyang 110869, China.*

## ABSTRACT

Diffusion models have demonstrated powerful data generation capabilities in various research fields such as image generation. However, in the field of vibration signal generation, the criteria for evaluating the quality of the generated signal are different from that of image generation and there is a fundamental difference between them. At present, there is no research on the ability of diffusion model to generate vibration signal. In this paper, a Time Series Diffusion Method (TSDM) is proposed for vibration signal generation, leveraging the foundational principles of diffusion models. The TSDM uses an improved U-net architecture with attention block, ResBlock and TimeEmbedding to effectively segment and extract features from one-dimensional time series data. It operates based on forward diffusion and reverse denoising processes for time-series generation. Experimental validation is conducted using single-frequency, multi-frequency datasets, and bearing fault datasets. The results show that TSDM can accurately generate the single-frequency and multi-frequency features in the time series and retain the basic frequency features for the diffusion generation results of the bearing fault series. It is also found that the original DDPM could not generate high quality vibration signals, but the improved U-net in TSDM, which applied the combination of attention block and ResBlock, could effectively improve the quality of vibration signal generation. Finally, TSDM is applied to the small sample fault diagnosis of three public bearing fault datasets, and the results show that the accuracy of small sample fault diagnosis of the three datasets is improved by 32.380%, 18.355% and 9.298% at most, respectively.

**Keywords:** diffusion model, time series, diffusion generation, small sample, fault diagnosis.

# 1. Introduction

In the field of rotational machinery fault diagnosis based on Machine Learning (ML), research often necessitates extensive training data to build ML models[1-3]. However, collecting a substantial volume of training data in practical engineering settings can be excessively time-consuming, expensive, or even infeasible[4]. Consequently, the challenge of fault diagnosis with small samples has garnered widespread attention among researchers[5,6]. The primary approach to address this issue is dataset expansion[7]. Presently, dataset expansion primarily relies on techniques such as interpolation to generate additional data from the small samples, forming an adequate training set for ML models[8,9]. Data generation methods encompass various data augmentation techniques [10,11], generative adversarial networks (GANs) [12-14], and Variational Auto-Encoder (VAE)[15-17]. Li et al. [45] proposed a data augmentation method based on diverse signal processing techniques, and the results indicated that with a sufficiently large number of generated samples, the diagnostic performance of fault diagnosis models improved. Ma et al. [46] proposed an enhanced version of traditional GANs known as Sparse Constraint GAN (SCGAN). SCGAN exhibited good convergence properties and effectively improved diagnosis accuracy. Wang et al. [47] presented an approach based on Sub-Pixel Convolutional Neural Networks (ESPCN), which could produce high-quality synthetic data and significantly improve the accuracy of rotational machinery fault diagnosis. Kingma et al.[15] proposed a general methodology based on Auto-Encoding method combined with variable lower bound to solve the hidden variables of Bayes graph model. VAE is a specific example of this methodology. Turinici[18] proposed a Radon Sobolev Variational Auto-Encoders (RS-VAE) by introducing a class of distances with built-in convexity to solve the shortcomings of convexity and fast evaluation in Wasserstein distance, slice Wasserstein distance, Jensen Shannon divergence, Kullback-Leibler divergence. In the field of sample generation, GANs exhibit superiority in sample quality. However, the training process is characterized by instability and lacks rigorous mathematical derivations. Consequently, improvements in GANs primarily focus on enhancing training stability. On the other hand, VAEs offer a mathematically rigorous foundation but struggle with generating high-quality samples. Hence, efforts to enhance VAEs concentrate on improving sample generation quality. The Diffusion model, in contrast, effectively addresses the shortcomings of both approaches, holding the potential to become a robust time series generation model.

Diffusion model is a new generation model that has developed rapidly in the field of Artificial Intelligence Generated Content (AIGC) in recent years[19]. It is called the new State of The Art (SOTA) model in the deep generation model[20]. The concept of diffusion model was proposed by Sohl-Dickstein et al.[21] in 2015, it was inspired by the diffusion movement in thermodynamics. For example, we drop a drop of red dye into a glass of pure water, the diffusion of dye molecules in the water is random. After a long enough time, the red dye will be evenly dispersed in the water, becoming a glass of red dye solution. If we record the diffusion trajectory of each red dye molecule and move it in the opposite direction, we can eventually get a drop of red dye and a cup of pure water again. This reverse movement process is the generative process. Suppose we process another cup of the same concentration of dye solution according to the recorded trajectory information. In that case, we will also theoretically get a drop of dye and a cup of pure water. Suppose we record the diffusion trajectory information of different concentrations of dye solutions. In that case, we can eventually achieve the reverse production of different concentrations of dye solutions into dye and pure water. This process of diffusion and reversal can be regarded as a diffusion model.

Diffusion model could only generate low-pixel images at first, but it began to be widely promoted in 2020. Berkeley et al.[22] proposed Denoising Diffusion Probabilistic Models (DDPM) for image generation, which surpassed Generative Adversarial Nets (GANs) in the authenticity, diversity and even aesthetic of the generated images, and the training process was more stable. In DDPM, U-net[23] is introduced to train and predict noise, significantly improving the diffusion model's diffusion generation ability. Since then, DDPM has demonstrated powerful capabilities in many fields[24]. In Computational Vision (CV), Saharia et al.[25] proposed a general conditional diffusion model for image-to-image translation, superior to GANs in four tasks: colourization, inpainting, uncropping, and JPEG decompression. Batzolis et al.[26] proved the superiority of the score-based diffusion model through theoretical analysis and introduced a multi-speed diffusion framework to improve the model, creating a benchmark for studying multi-speed diffusion. Yang et al.[27] proposed neural video coding algorithms presented various architectures that achieve state-of-the-art performance in compressing high-resolution videos and delved into their trade-offs and variations. Rombach et al. [28] proposed a novel model that combines a diffusion model with highly effective pretrained autoencoders. This integration enabled the training of diffusion models even with constrained computational resources while maintaining their quality and flexibility. In contrast to prior research, training diffusion models on such a representation allowed for achieving a nearly optimal balance between complexity reduction and detail preservation, significantly enhancing visual fidelity. Yang et al. [29] proposed an autoregressive, end-to-end optimized video diffusion model, drawing inspiration from recent advancements in neural video compression. This model sequentially produces forthcoming frames by refining a deterministic next-frame prediction by integrating a stochastic residual generated via an inverse diffusion process. Furthermore, owing to its formidable generative capabilities, diffusion models have made substantial strides in various domains, including Natural Language Processing (NLP) [30-32], Waveform Signal Processing [33,34], Molecular Graph Modeling [35-38], and Adversarial Purification [39-41].

Diffusion model has shown high quality generation ability in time series. However, the field of application is mainly focused on simple time series, such as weather trends, audio signals[44] and ECG signals. These signals have strong regularity, single composition, and less interference components in them. For instance, the vibrations in the ECG signal are so concentrated and specific that experienced researchers can obtain useful information by simply analyzing them, so it is easy to obtain the pattern through deep learning methods. In contrast, the generation of vibration signals is a more complex task. Vibration signal is usually obtained by collecting the vibration of rotating machinery, which has complex structure, high rotating speed and many interfering factors in operation. The vibration signal is usually obtained by collecting the vibration of rotating machinery, which has complex structure, high rotating speed and many interfering factors in operation. This leads to a more complex composition of vibration signals. Moreover, in the same time period, the data of vibration signal changes more greatly and the law is more complex. Therefore, vibration signal is more complicated to summarize, and the generation of vibration signal is more difficult. At present, no method based on diffusion model for vibration signal generation of rotating machinery has been proposed.

*To address the above issues further, we propose a Time Series Diffusion Method (TSDM) for vibration signal generation, leveraging the data generation capabilities of DDPM. The TSDM enhances the improved U-net architecture with attention block, ResBlock and TimeEmbedding to enable segmentation and feature extraction of one-dimensional time series data, and it is founded on forward diffusion and reverse denoising processes for time series generation. TimeEmbedding enables U-net to record the times of noise additions and denoising, which will greatly improve the efficiency of the training network. Through TSDM-based generation experiments on single-frequency, multi-frequency, and bearing datasets. The accuracy and effectiveness of the TSDM generation results are validated, and the generated results are significantly better than existing methods. The test results also show that the original DDPM cannot generate high-quality vibration*

*signals, but the improved U-net in TSDM, which uses the combination of attention block and ResBlock, can effectively improve the quality of vibration signal generation. Finally, the TSDM is applied to small sample fault diagnosis on three public bearing fault datasets, demonstrating that its application significantly improves the accuracy of small sample fault diagnosis, and the effect is better than other methods.*

## 2. Basic Theory

Denoising Diffusion Probabilistic Models (DDPM)[22] are based on the diffusion model, including forward diffusion, reverse denoising processes and model optimization. The specific principle is as follows.

2.1 Forward Diffusion Process

The forward diffusion process is the process of gradually adding Gaussian noise to the data until it ultimately becomes random noisy data. For the raw data $x_0$ that will undergo $T$-step diffusion, the result $x_t$ obtained from each diffusion step is obtained by adding Gaussian noise to the previous step data $x_{t-1}$, described in Eq.(1).

$$q\left(x_t \middle| x_{t-1}\right) = \mathbb{N}\left(x_t \middle| \sqrt{1-\beta_t}x_{t-1}, \beta_t \mathbf{I}\right) \tag{1}$$

where $\{\beta_t\}_{t=1}^{T}$ is the variance of Gaussian distribution noise at each step; $q(x_t)$ is the probability distribution of the data $x_t$. As step $t$ increases, the variance $\beta_t$ needs to be taken larger, but it needs to satisfy as follow:

$$0 < \beta_1 < \beta_2 \cdots \beta_{T-1} < \beta_T < 1 \tag{2}$$

If the diffusion step $T$ is large enough, the result data will lose its original information and become random noise data. The entire diffusion process is a Markov chain from $t$=1 to $t$=$T$:

$$q\left(x_{1:T} \middle| x_0\right) = \prod_{t=1}^{T} q\left(x_t \middle| x_{t-1}\right) \tag{3}$$

The diffusion process is often fixed by using a pre-defined variance schedule. An essential feature of the forward diffusion process is that it can directly sample $x_t$ at any step $t$ based on the original data $x_0$: $x_t \sim q(x_t \mid x_0)$. If $\alpha_t = 1-\beta_t$ and $\bar{\alpha}_t = \prod_{i=1}^{t} \alpha_i$ are defined, then through the reparamazation, the diffusion process can be expressed as follows:

$$\begin{aligned} x_t &= \sqrt{\alpha_t}x_{t-1} + \sqrt{1-\alpha_t}\varepsilon_{t-1} \\ &= \sqrt{\alpha_t}\left(\sqrt{\alpha_{t-1}}x_{t-2} + \sqrt{1-\alpha_{t-1}}\varepsilon_{t-2}\right) + \sqrt{1-\alpha_t}\varepsilon_{t-1} \\ &= \sqrt{\alpha_t \alpha_{t-1}}x_{t-2} + \sqrt{1-\alpha_t \alpha_{t-1}}\bar{\varepsilon}_{t-2} \\ &= \cdots \\ &= \sqrt{\bar{\alpha}_t}x_0 + \sqrt{1-\bar{\alpha}_t}\varepsilon \end{aligned} \tag{4}$$

where $\varepsilon_{t-1},\varepsilon_{t-1},\cdots\sim\mathbb{N}(0,\mathbf{I})$, and $\bar{\varepsilon}_{t-2}$ merges two Gaussians. $\{\alpha_t\}_{t=1}^{T}$ can be called the noise schedule. $\bar{\alpha}_t=\prod_{i=1}^{t}\alpha_i$ is a hyperparameter set with a noise schedule. Then, the diffusion process can be expressed as follows:

$$q\left(x_t|x_0\right)=\mathbb{N}\left(x_t;\sqrt{\bar{\alpha}_t}x_0,\left(1-\bar{\alpha}_t\right)\mathbf{I}\right) \tag{5}$$

The above is the entire process of the forward diffusion progress. $x_t$ can be seen as a linear combination of the original data $x_0$ and random noise $\varepsilon$, where $\sqrt{\bar{\alpha}_t}$ and $\sqrt{1-\bar{\alpha}_t}$ are the combination coefficients. Adjusting parameter $\bar{\alpha}_T$ to change the results generated by diffusion is more direct than variance $\beta_t$. For example, if $\bar{\alpha}_T$ is set to a value close to 0, the resulting data is closer to Gaussian noise; If $\bar{\alpha}_T$ is set to a value close to $T$, the resulting data is closer to the original data.

2.2 Reverse Denoising Process

The reverse denoising process is based on the true distribution $q(x_{t-1} \mid x_t)$ of each step, gradually denoising from a random noise $x_T\sim\mathbb{N}(0,\mathbf{I})$, and ultimately generating the target data. Use a neural network to learn the distribution $q(x_{t-1} \mid x_t)$ of the entire training sample and obtain the parameterized distribution $p_\theta(x_{t-1}\mid x_t)$ of the neural network. The reverse process is also defined as a Markov chain. $p_\theta$ can be expressed as follows:

$$p_\theta\left(x_{t-1}|x_t\right)=\mathbb{N}\left(x_{t-1};\mu_\theta\left(x_t,t\right),\sum_\theta\left(x_t,t\right)\right) \tag{6}$$

$$p_\theta\left(x_{0:T}\right)=p\left(x_T\right)\prod_{t=1}^{T}p_\theta\left(x_{t-1}|x_t\right) \tag{7}$$

where $\theta$ is a parameter of the neural network. $p(x_T)=\mathbb{N}(x_T;0,\mathbf{I})$ a random Gaussian noise, $p_\theta\left(x_{t-1}|x_t\right)$ is a parameterized Gaussian distribution that requires the training network to calculate the mean $\mu_\theta(x_t,t)$ and variance $\sum_\theta(x_t,t)$.

3.3 Model Optimization

In the reverse denoising process, the true distribution $q(x_{t-1} \mid x_t)$ is approximated to the parameterized distribution $p_\theta(x_{t-1}\mid x_t)$ of the neural network. The optimization goal of TSDM is to make $p_\theta(x_{t-1}\mid x_t)$ as close to $q(x_{t-1} \mid x_t)$ as possible. This can be translated into finding the minimum KL divergence[20] of two joint distributions for two distributions, which can be defined as the loss function $L_t$:

$$L_t=D_{KL}\left(q\left(x_{1:T}|x_0\right)\middle\|p_\theta\left(x_{1:T}|x_0\right)\right) \tag{8}$$

The mean of $p_\theta(x_{t-1}\mid x_t)$ and $q(x_{t-1} \mid x_t)$ can be written as follow:

$$\mu_\theta\left(x_t,t\right)=\frac{1}{\sqrt{\alpha_t}}\left(x_t-\frac{1-\alpha_t}{\sqrt{1-\alpha_t}}\varepsilon_\theta\left(x_t,t\right)\right) \tag{9}$$

$$\tilde{\mu}_t(x_t,t)=\frac{1}{\sqrt{\alpha_t}}\left(x_t-\frac{1-\alpha_t}{\sqrt{1-\alpha_t}}\varepsilon_t(x_t,t)\right) \tag{10}$$

The loss function $L_{t-1}$ of step $t$-1 can be written as:

$$\begin{aligned}
L_{t-1}&=\mathrm{E}_{q(x_t|x_0)}\left[\frac{1}{2\sigma_t^2}\left\|\tilde{\mu}_t(x_t,x_0)-\mu_\theta(x_t,t)\right\|^2\right]\\
&=\mathrm{E}_{x_0}\left(\mathrm{E}_{q(x_t|x_0)}\left[\frac{1}{2\sigma_t^2}\left\|\tilde{\mu}_t(x_t,x_0)-\mu_\theta(x_t,t)\right\|^2\right]\right)\\
&=\mathrm{E}_{x_0,\varepsilon\sim\mathbb{N}(0,\mathbf{I})}\left[\frac{1}{2\sigma_t^2}\left\|\tilde{\mu}_t\left(x_t,\frac{1}{\sqrt{\tilde{\alpha}_t}}\left(x_t-\sqrt{1-\tilde{\alpha}_t}\varepsilon\right)\right)-\mu_\theta(x_t,t)\right\|^2\right]\\
&=\mathrm{E}_{x_0,\varepsilon\sim\mathbb{N}(0,\mathbf{I})}\left[\frac{1}{\sqrt{2\sigma_t^2}}\left\|\frac{1}{\sqrt{\alpha_t}}\left(x_t-\frac{\beta_t}{\sqrt{1-\tilde{\alpha}_t}}\varepsilon\right)-\mu_\theta(x_t,t)\right\|^2\right]\\
&=\mathrm{E}_{x_0,\varepsilon\sim\mathbb{N}(0,\mathbf{I})}\left[\frac{1}{2\sigma_t^2\alpha_t(1-\tilde{\alpha}_t)}\left\|\varepsilon-\varepsilon_\theta\left(\sqrt{\tilde{\alpha}_t}x_0+\sqrt{1-\tilde{\alpha}_t}\varepsilon,t\right)\right\|^2\right]
\end{aligned} \tag{11}$$

where $x_t$ represents the $x_t(x_0, \varepsilon)$ obtained by adding noise $\varepsilon$ to original data $x_0$. $\varepsilon_\theta$ is a fitting function based on neural networks, which means that the model has switched from the original predicted mean to the predicted noise $\varepsilon$.

By removing the weight coefficients of the target loss function, a further simplified result can be obtained as follows[22]:

$$L^{simple}(\theta)=\mathrm{E}_{t,x_0,\varepsilon\sim\mathbb{N}(0,\mathbf{I})}\left[\left\|\varepsilon-\varepsilon_\theta\left(\sqrt{\tilde{\alpha}_t}x_0+\sqrt{1-\tilde{\alpha}_t}\varepsilon,t\right)\right\|^2\right] \tag{12}$$

# 3. Time Series Diffusion Method

## 3.1 Improved U-net architecture

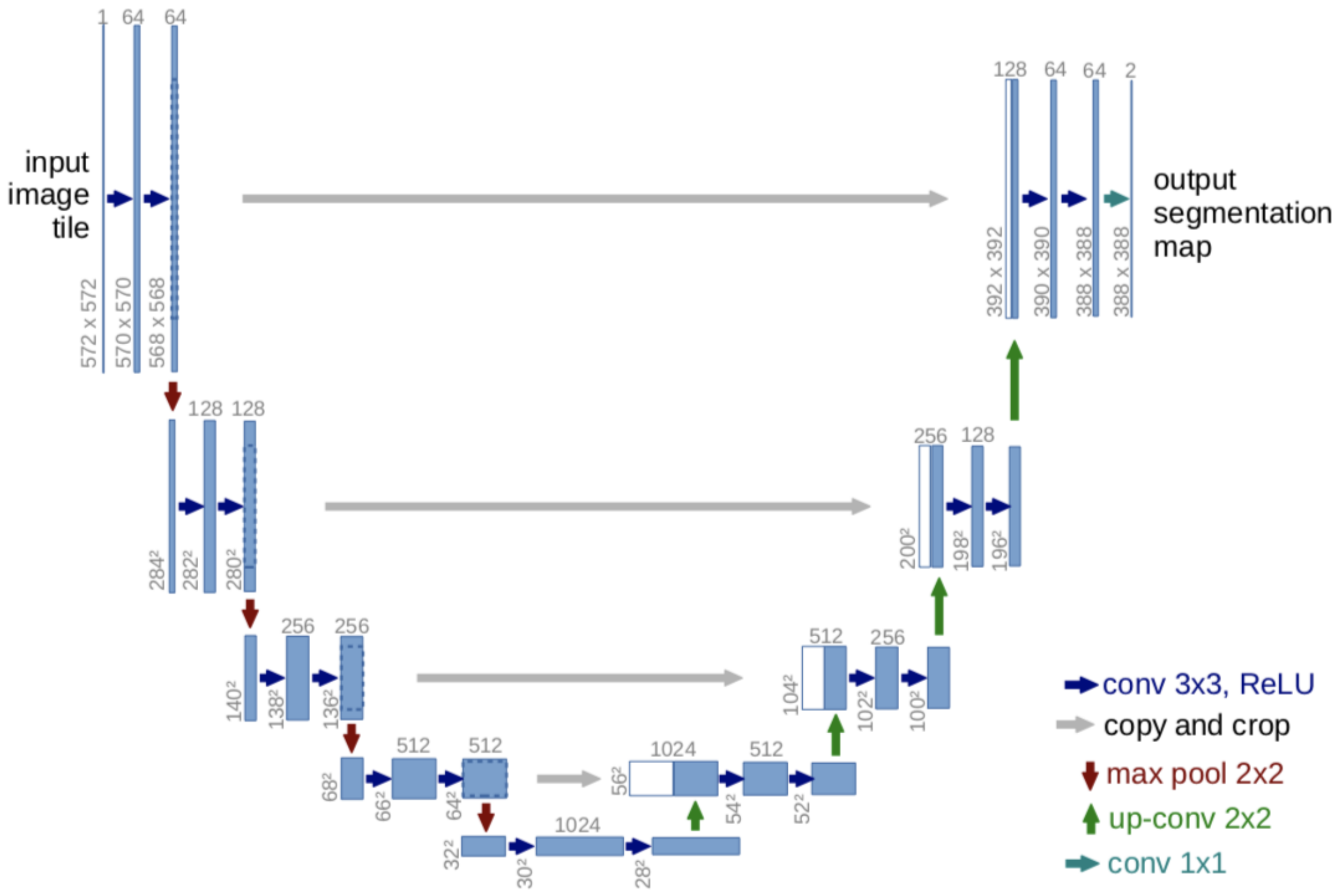

Fig. 1 Original U-net architecture[23].

U-net [23] is widely used in the field of semantic segmentation in image processing and machine vision. The original U-net architecture is shown in Fig. 1. In the TSDM proposed in this study, TimeEmbedding, ResBlock and AttnBlock are introduced to improve U-net to realize the noise prediction mechanism. The improved U-net architecture in this study is asymmetric, as shown in Fig. 2.

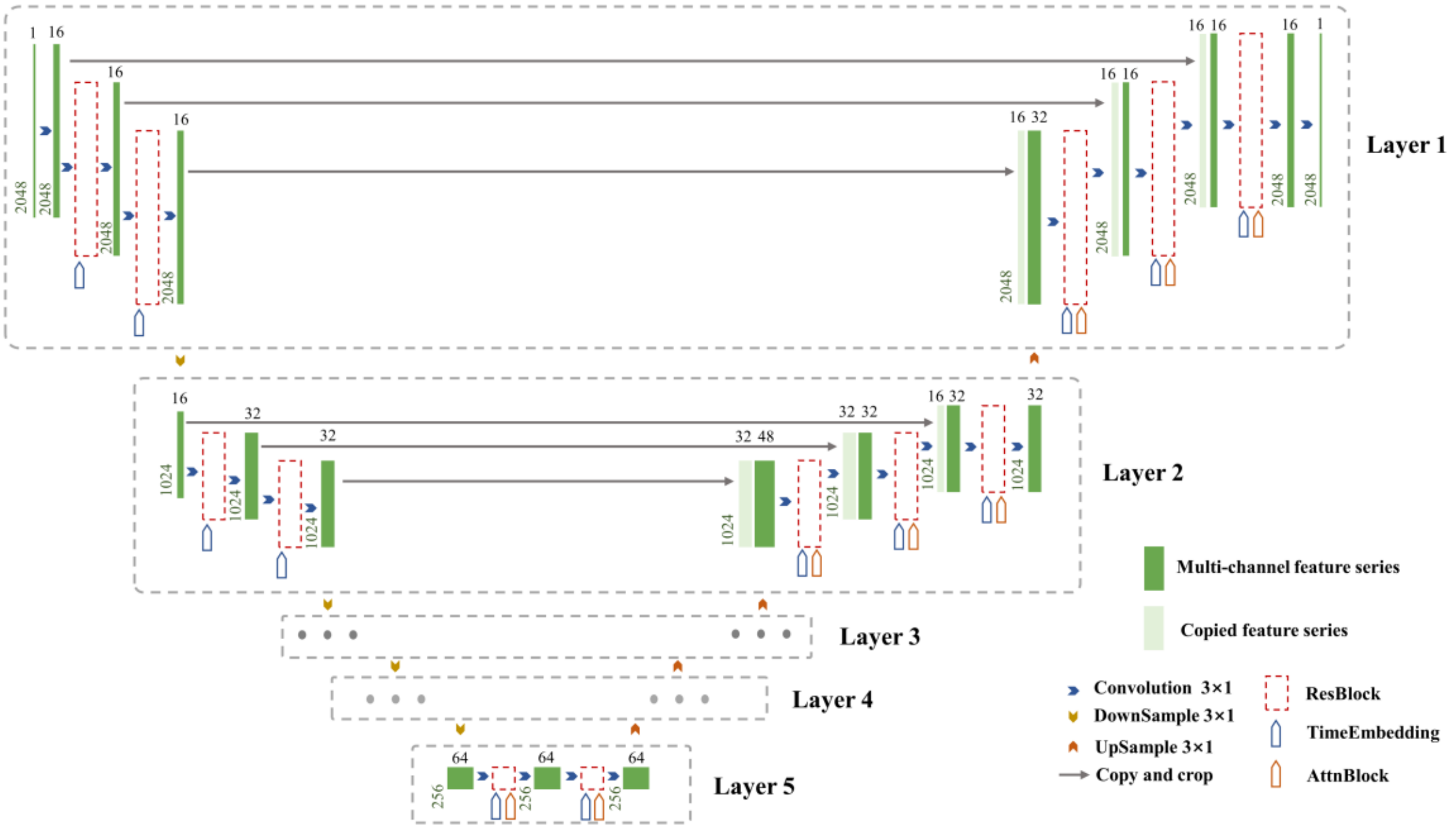

Fig. 2 The improved U-net architecture of the Time Series Diffusion Method

In the down-sampling process, the feature series enters the DownSample block after four convolutions and two ResBlock. The DownSample block can save practical information and reduce the dimension of features to avoid overfitting. In the middle sampling stage, the feature series enters the UpSample block after four convolutions and two ResBlock. The AttnBlock is added to each ResBlock to retain features. In the up-sampling stage, the feature series enters the UpSample block after six convolutions and three ResBlock. Each feature series in the down-sampling process is copied and concatenated in the up-sampling process to achieve the retention of the same dimensional features, which is conducive to network optimization. The AttnBlock is applied in the 3rd ResBlock to achieve better learning of features and increase the global modelling ability of the network. The TimeEmbedding is fused with feature series in each ResBlock for model prediction and can implement U-net model sharing.

**3.2 TSDM architecture**

Based on the forward diffusion and reverse denoising processes in the Basic Theory, combined with the improved U-net and the loss function used to optimize the network, the architecture of TSDM is shown in Fig. 3 to Fig. 5. The training diagram of TSDM is demonstrated in Fig. 3.

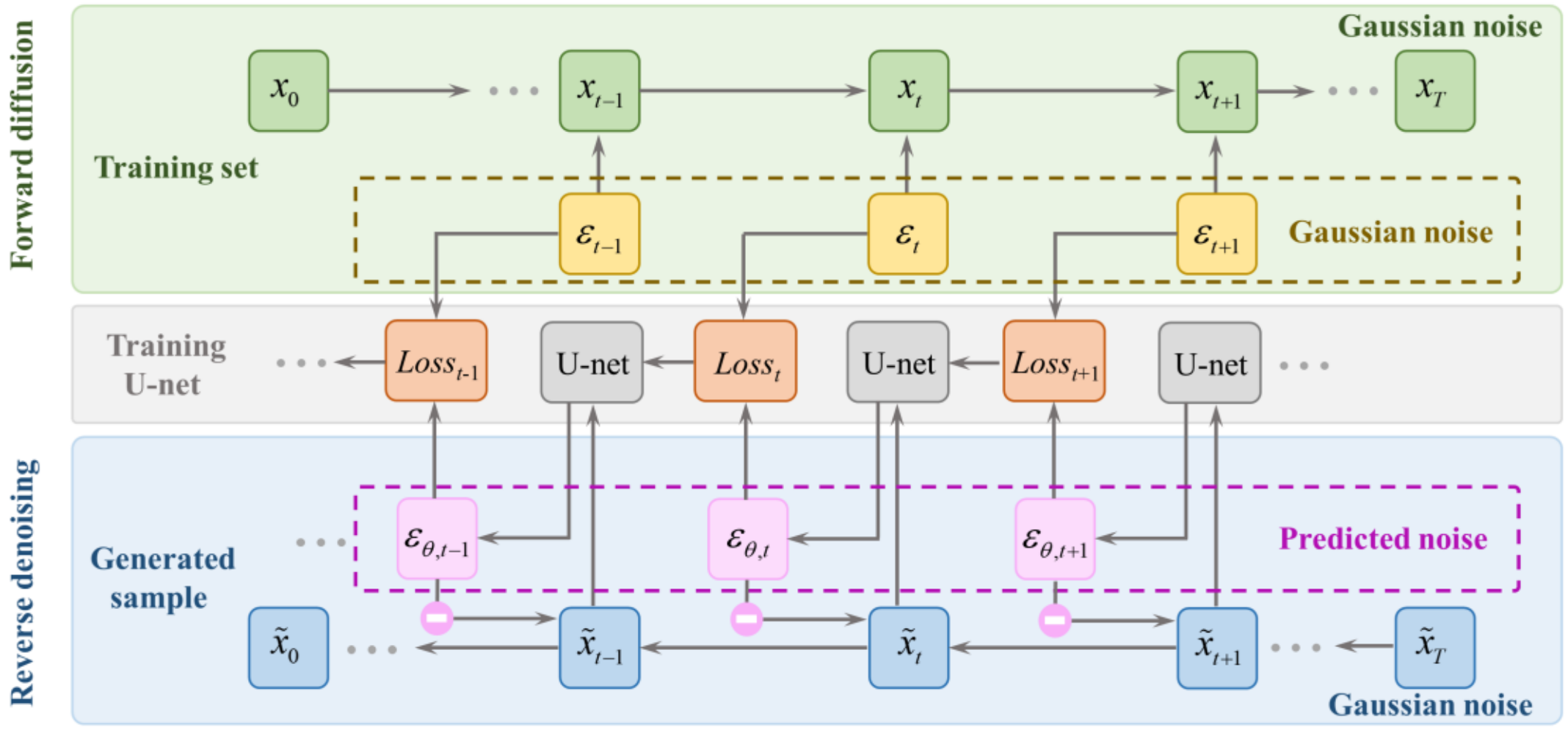

Fig. 3 The training process of TSDM.

The training of TSDM is essentially the training of the U-net neural network in the model. In the forward diffusion process, Gaussian noise $\varepsilon_t$ is added to the training sample $x_t$ at each step $t$, and finally, $x_T$ is generated through $T$-step diffusion, which is almost Gaussian noise. In the reverse denoising process, for each $\tilde{x}_t$, it is input into the U-net neural network to predict the denoising noise $\varepsilon_{\theta,t}$. The predicted noise $\varepsilon_{\theta,t}$ and the Gaussian noise $\varepsilon_t$ added in the forward diffusion process (training process) are substituted into the loss function formula to update the U-net model and realize the optimization of the model.

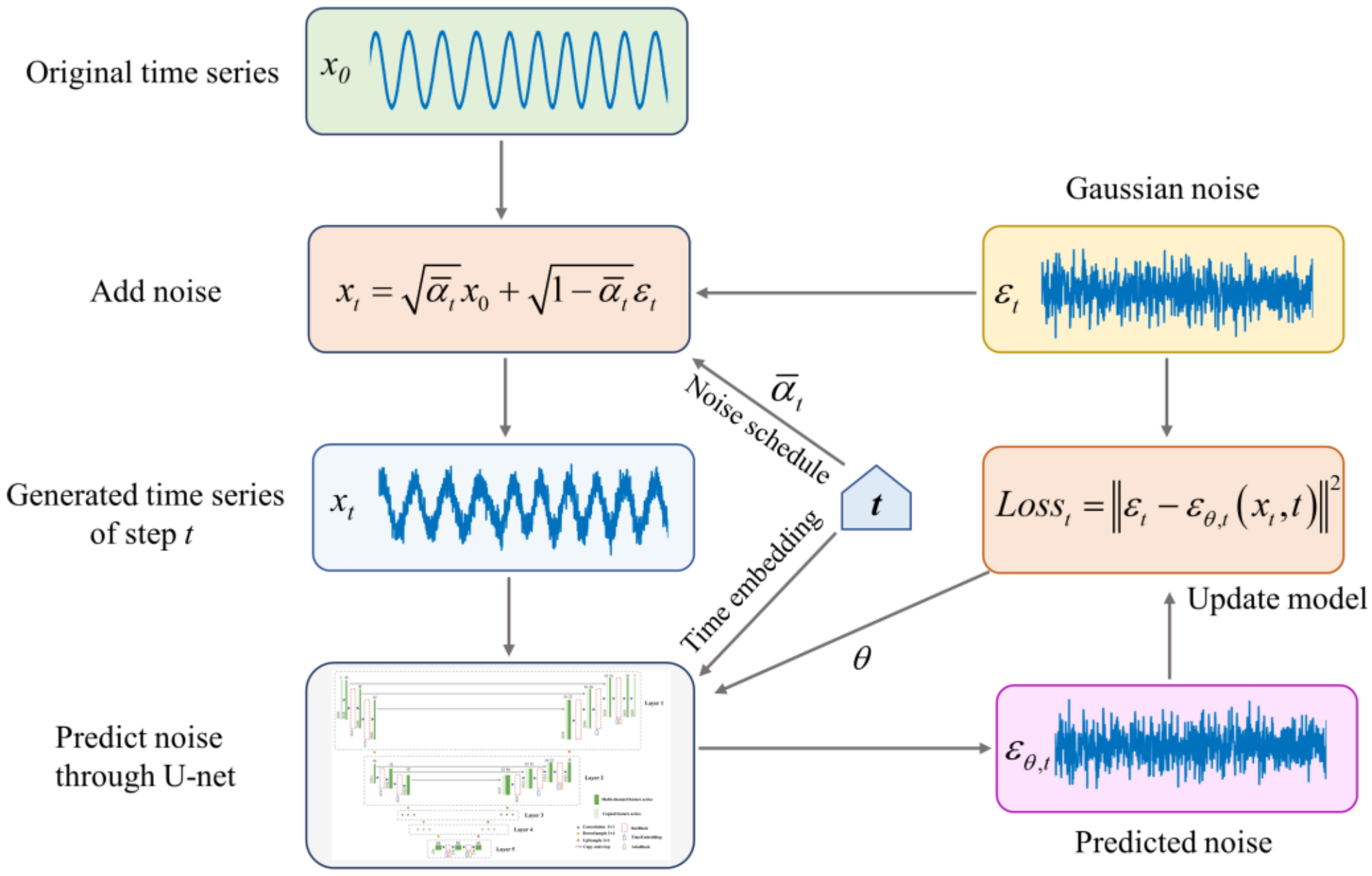

Fig. 4 Role of TimeEmbedding in the improved U-net training of TSDM.

Theoretically, TSDM needs to train the improved U-net at each $t \in [0, T]$, but the value of $T$ is usually greater than 1000, which leads to the slow training of U-net. Therefore, TSDM applies TimeEmbedding to optimize the training process, as shown in Fig. 4. The optimization is actually carried out through a random time step $t$, rather than through each time step $t$. The noise schedule is generated according to the time step $t$, and the generated time series with added noise is trained through U-net. TimeEmbedding is used to record and share the time step $t$ during the training process. The loss function between the noise $\varepsilon_{\theta,t}$ predicted by u-net and the added Gaussian noise $\varepsilon_t$ is calculated, and the neural network parameter $\theta$ is updated. Next, take another random time step $t$ for the next cycle until the end of the optimization process. At the end of the U-net optimization process, TSDM has also completed training and can generate time series by diffusion. The schematic diagram and diffusion generation process of TSDM is shown in Fig. 5.

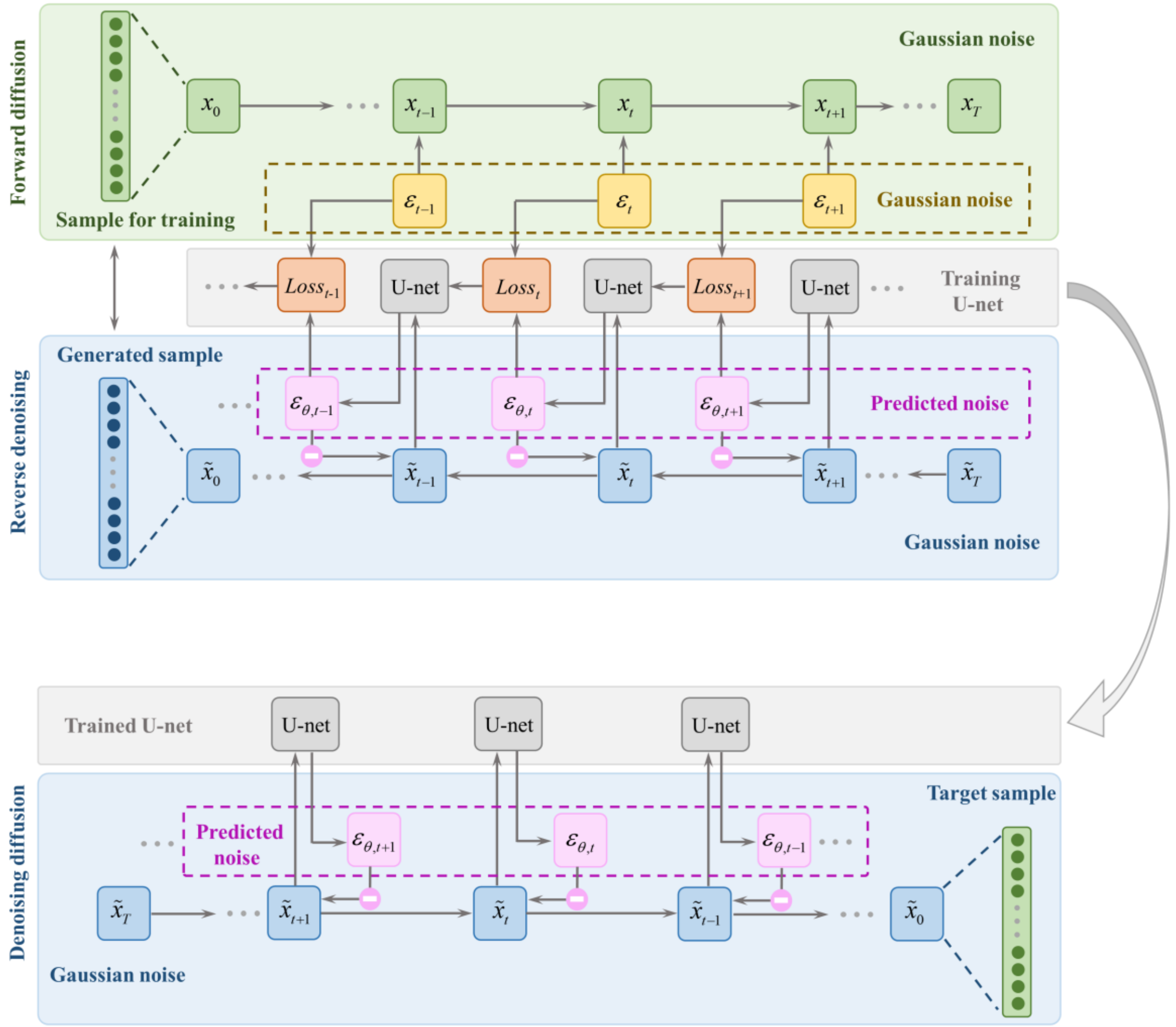

Fig. 5 Schematic diagram and the process of diffusion generation

The diffusion generation process is to denoise the Gaussian noise sample $\tilde{x}_T$ layer by layer. The final generated target sample is determined by the noise $\varepsilon_{\theta,t}$ denoised in each step $t$, and the noise $\varepsilon_{\theta,t}$ is predicted by the trained U-net. Finally, the target time series $\tilde{x}_0$ is generated, it has the characteristics of the training set and contains new random features, which makes the generated results expand the training set samples.

# 4. Experimental Results

In this section, taking the artificially constructed time series datasets and the published bearing fault datasets as examples, the effectiveness of the time series diffusion method proposed in this paper is tested by comparing the feature similarity between the generated series and the original series. The datasets used include single-frequency time series, multi-frequency time series and XJTU[52] bearing fault datasets.

## 4.1 Single-frequency Time Series

A single-frequency time series dataset is constructed by trigonometric function. The construction method is as follows:

$$x_n(\varphi) = \sin(2\pi k_1 \varphi + b_n) \quad (13)$$

where $k_1$ is the preset frequency; $b_n$ is a random number between 0 and 2π, used to make the phase difference between time series and avoid overfitting between data.

A single-frequency time series dataset of 10Hz is built according to Eq.(13), and the dataset size is [200,2048], which contains 200 time series with a length of 2048. Partial time series in the single-frequency dataset are shown in Fig. 6.

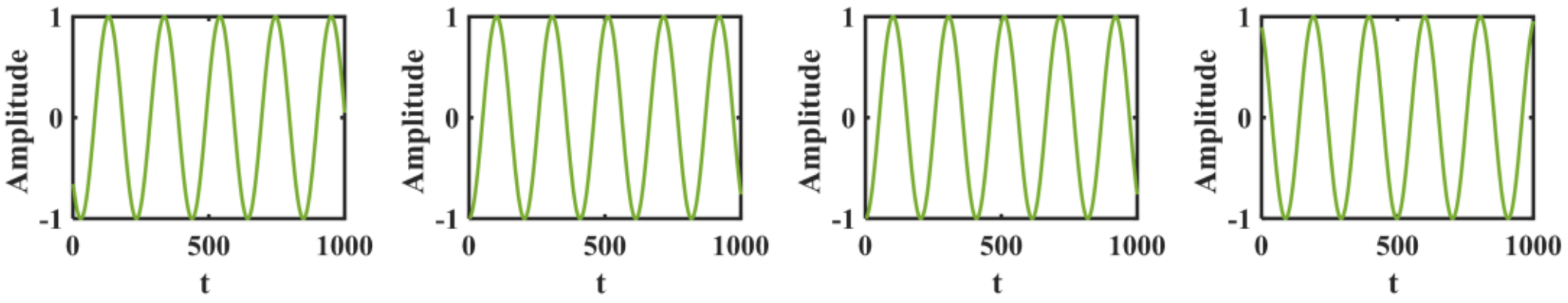

(a) (b) (c) (d)

Fig. 6 Partial time series in the 10Hz single-frequency dataset.

During the forward diffusion and training process, the batch size is set to 10, and the TSDM is trained over 200 epochs to realize denoising generation. The number of noise diffusion and denoising layers T is set to 3000. Based on the trained model, 40 target time series are generated, partial results are shown in Fig. 7, and the corresponding frequency spectrum is shown in Fig. 8.

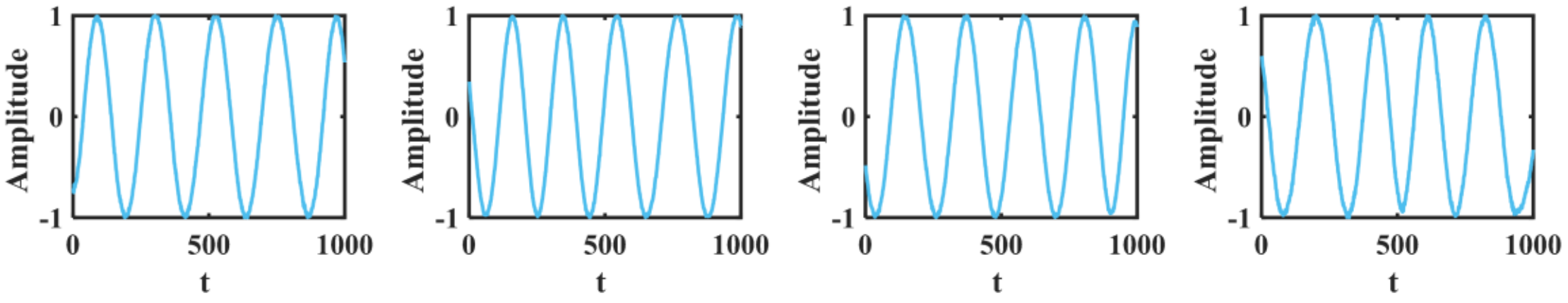

(a) (b) (c) (d)

Fig. 7 Generation results of 10Hz single frequency dataset.

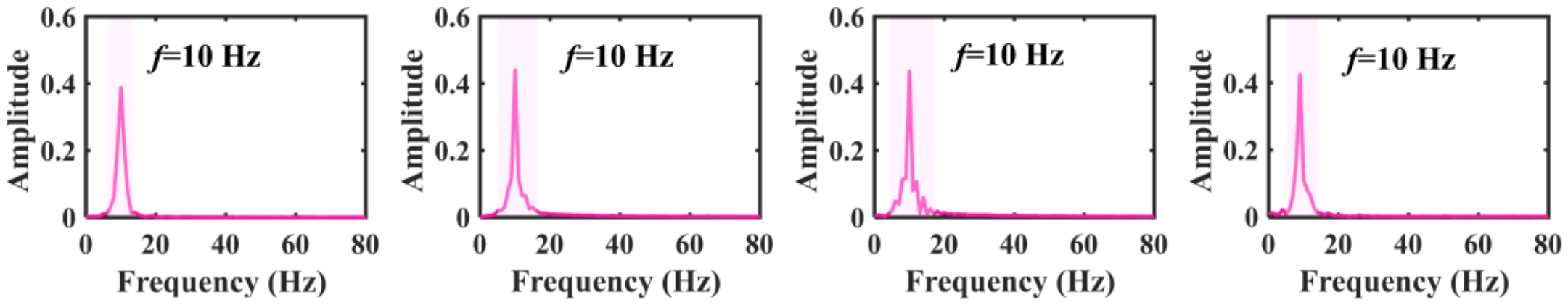

(a) (b) (c) (d)

Fig. 8 Corresponding frequency spectrum of generation results of 10Hz single frequency dataset.

In the diffusion generation results of 10Hz single frequency dataset, it can be seen in Fig. 7 that the time series generated by diffusion show a standard form of trigonometric function, and the periodicities are consistent. Fig. 8 shows the corresponding frequency spectrum of generation results, it can be seen that the characteristic frequency of 10Hz is well preserved after diffusion generation, which reflects the accuracy of generated results. However, there are differences in the bandwidth of the main peak of frequency spectrums, which is a manifestation of the randomness of the generated results. It means the generated results of the TSDM have certain randomness while retaining the main characteristics. It also shows that the TSDM can generate diversified target times series rather than simply copying training samples.

Summarize the frequency spectrums of 40 time series generated and draw the box plot as shown in Fig. 9. It can be seen from the box plot that the peak values of the frequency component of the generated time series are mainly around the characteristic frequency. However, the amplitude fluctuation of the peak is relatively large compared with other frequency positions, and it can be seen that the bandwidth of the average spectrum peak is significantly wider than that of a single sample. The uncertainty of the amplitude and bandwidth of the resulting spectrum peak reflects the creativity of the TSDM.

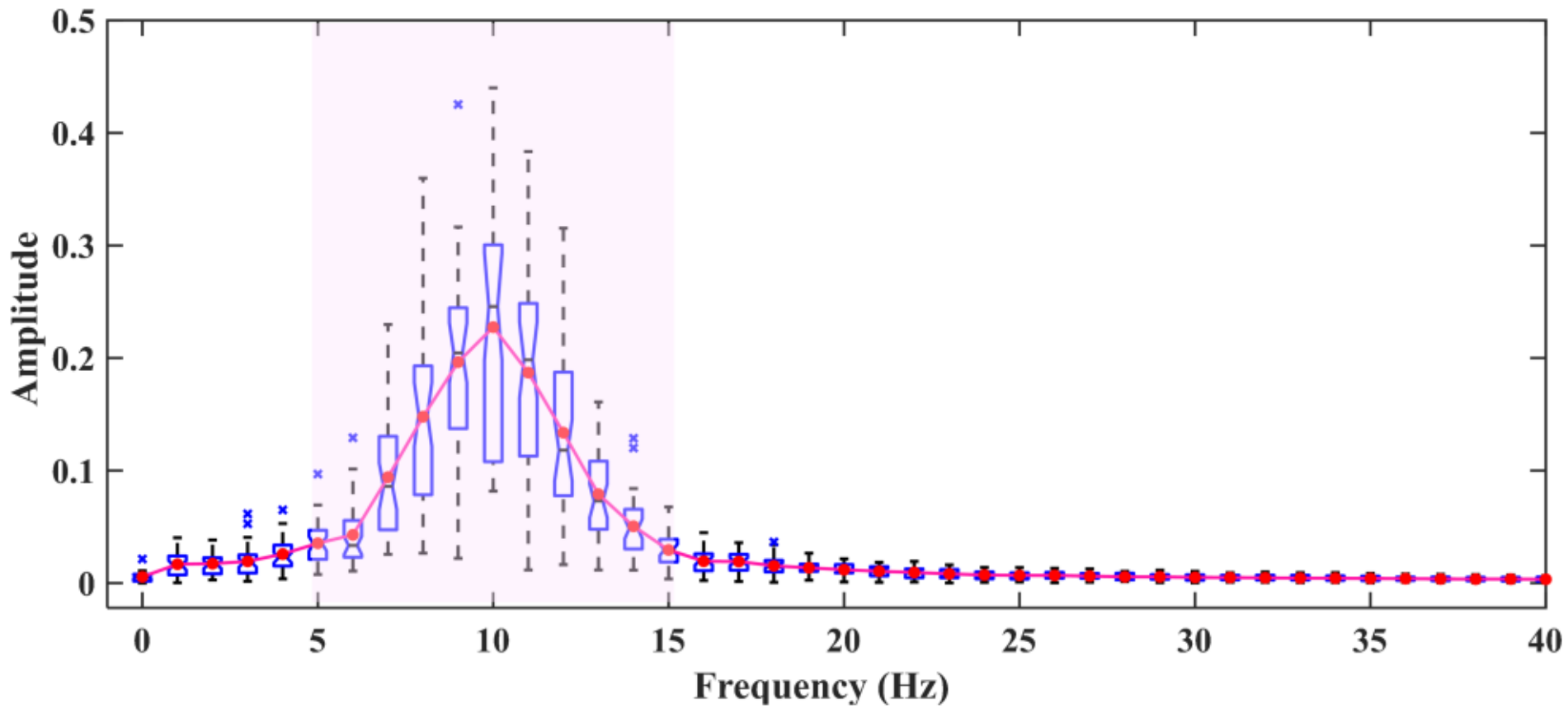

Fig. 9 Box plot of generation results of 10Hz single frequency dataset

Taking the generated result in Fig. 7(a) as an example, draw the process of gradually denoising it from random noise to generating a single-frequency trigonometric function, as shown in Fig. 10. From the denoising generation process, it can be seen that with the increase of denoising times *t*, the random noise first gradually forms the contour of the target sequence. For example, when *t*=2550, a rough outline appears, and when *t*=2850, the shape is so apparent that the periodicity of the target series can be seen. It can also be seen from the corresponding frequency spectrum that with the increase of denoising times, the corresponding peak of the characteristic frequency of the target series gradually appears and increases. In the U-net architecture of TSDM, because U-net is a shared parameter, the role of TimeEmbedding is to let the model form the general outline of the series and learn the critical feature information when it is close to generating the target series. This dramatically improves the efficiency of TSDM generation.

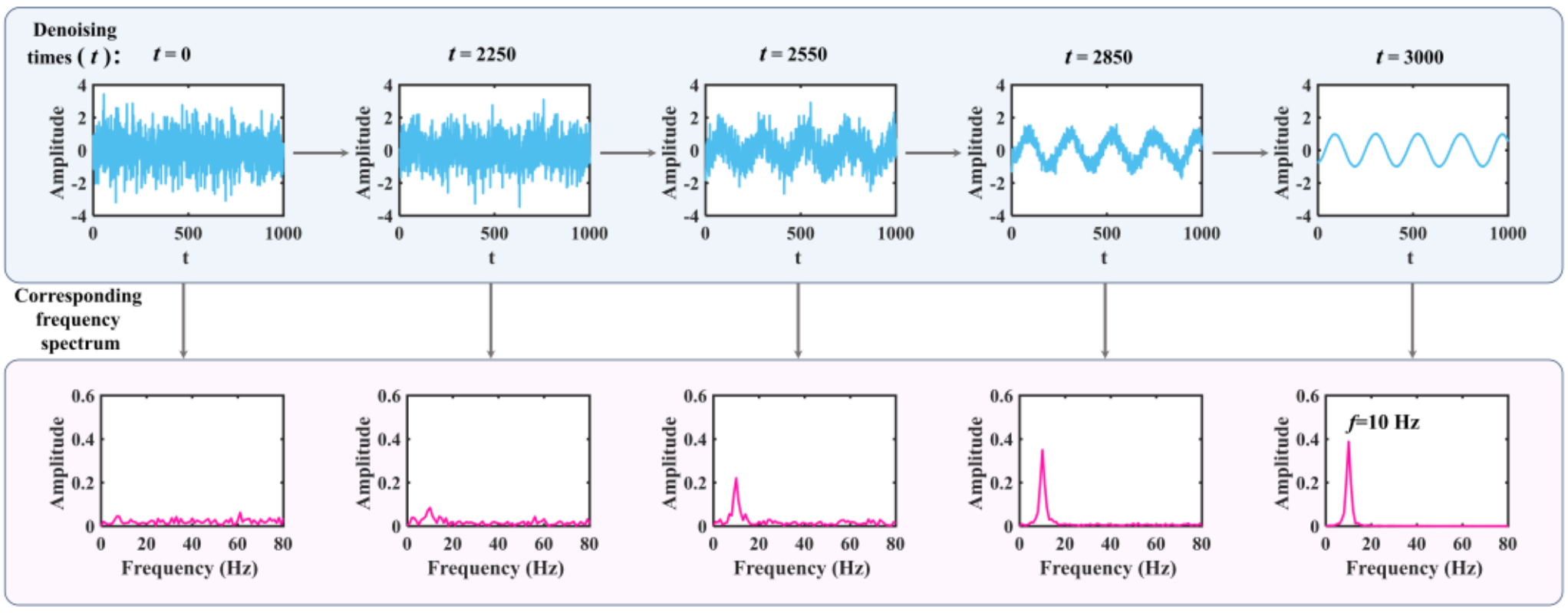

Fig. 10 Denoising generation process of TSDM for a single-frequency data example.

For the quality of time series and vibration signal generation cannot be simply evaluated with labels, we are more concerned about the consistency of the generated signal frequency with the original signal frequency. To evaluate the generation quality of time series and vibration signals by different methods, the Variance Frequency (VF) is introduced:

$$VF\left(p,n\right)=\sum_{p=1}^{P}\frac{\sum_{n=1}^{N}\left|f_p(n)-\hat{f}_p\right|^2}{\sum_{n=1}^{N}f_p(n)} \tag{14}$$

where $\hat{f}_p$ is the target frequency of the generated signal. $f_p(n)$ is the actual frequency of the generated signal. $VF\left(p,n\right)$ means that there are $n$ sets of generated signals, each with $p$ target frequencies. $VF$ value reflects the consistency between the corresponding frequency spectrum of the generated time series and the frequency of the target series. The smaller the VF value is, the higher the consistency and the quality of the generated time series is. The higher the VF value is, the lower the quality of the generated time series is.

For the generation of single frequency time series, the TSDM proposed in this paper is compared with the existing time series generation methods. The comparison focuses on the waveform coincidence degree and VF value between the generated time series and the target time series. The results are shown in Fig. 11 and Tab. 1.

From the waveform results of the generated time series in Fig. 11, it can be seen that for each time series generated by VQ-VAE[42] and TimeGAN[43], only a part of the waveform can be consistent with the target waveform, but the high-quality generation of the whole time series cannot be achieved. For the Diffwave[44] method based on the diffusion model, the waveform of the generated time series contains many high-frequency components. Although it is effective in the generation of audio signals, it does not work well for the time series of single-frequency and whole period. The proposed TSDM excellently realizes the generation of single-frequency whole periodic time series, and the waveform is in good agreement with the target time series, with only slight error. From the *VF* values of the four methods results in Tab. 1, it can also be seen that the *VF* values of the times series generated by the TSDM method are significantly lower than those of the other three methods, indicating that the accuracy of the TSDM generated results is higher and TSDM can preserve frequency characteristics better.

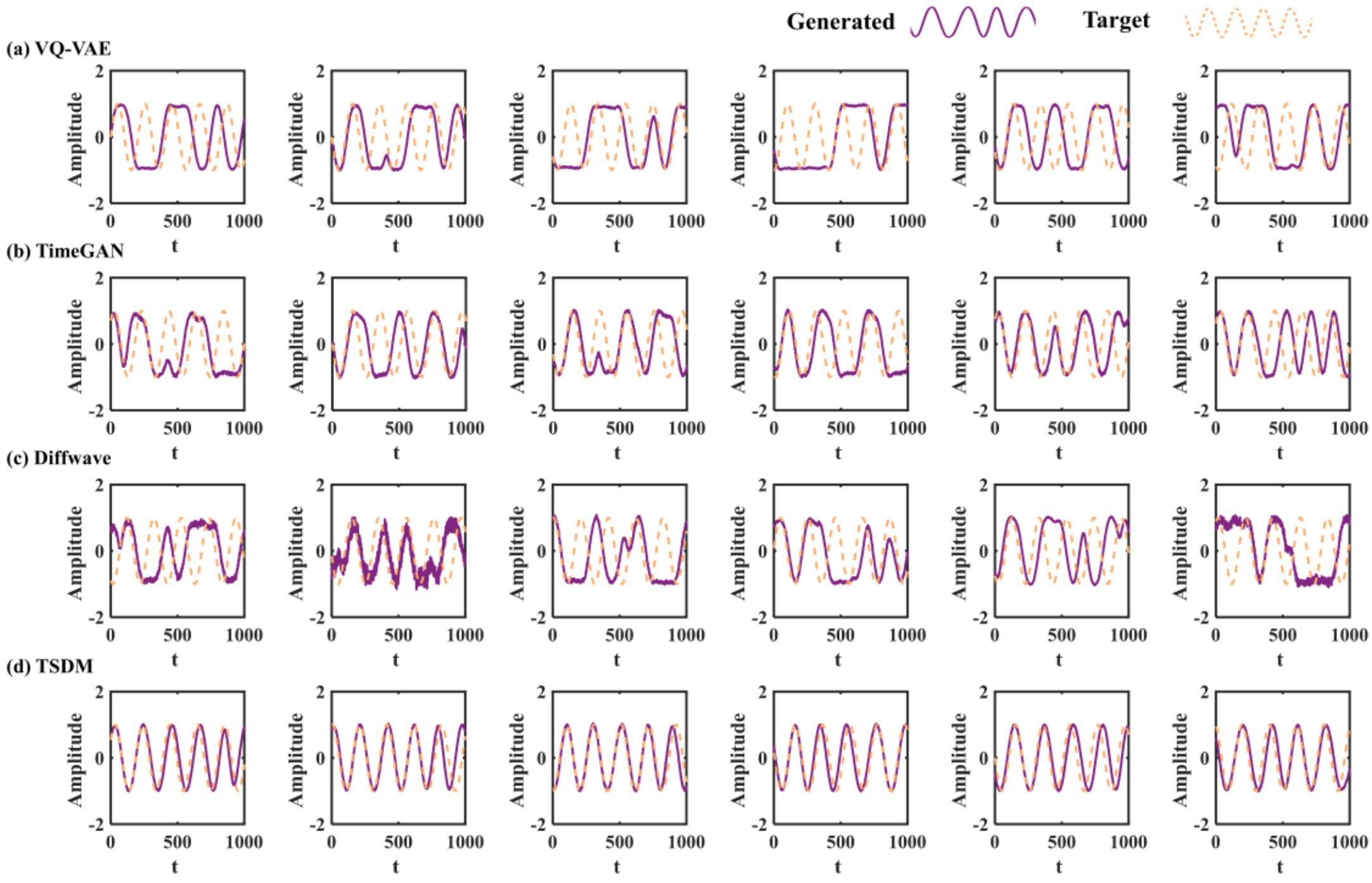

Fig. 11 Waveform quality of the time series generated by (a) VQ-VAE. (b) TimeGAN. (c) Diffwave. (d) proposed TSDM.

Tab. 1 *VF* value of the time series generated by different methods.

| | VQ-VAE | TimeGAN | Diffwave | TSDM |
|---|---|---|---|---|
| Variance Frequency | 1.9946 | 1.4633 | 1.6462 | **0.2658** |

### 4.2 Multi-frequency Time Series

A multi-frequency time series dataset is also constructed by trigonometric function. The construction method is as follows:

$$x_n(\varphi) = \sin(2\pi k_1 \varphi + b_{1n}) + \sin(2\pi k_2 \varphi + b_{2n}) + \cdots + \sin(2\pi k_m \varphi + b_{mn}) \qquad (15)$$

where $k_1$, $k_2 \cdots k_m$ is the preset frequencies; $b_{1n}$, $b_{2n} \cdots b_{mn}$ is the random number between 0 and $2\pi$, which is used to make the phase difference between time series with the same frequency and avoid overfitting.

In this study, three time series with different frequencies are combined into a multi-frequency time series according to Eq.(14), where $k_1$=88 $k_2$=222 $k_3$=333. The dataset size is [200,2048], containing 200 time series with a length of 2048. Partial time series in the multi-frequency dataset are shown in Fig. 12.

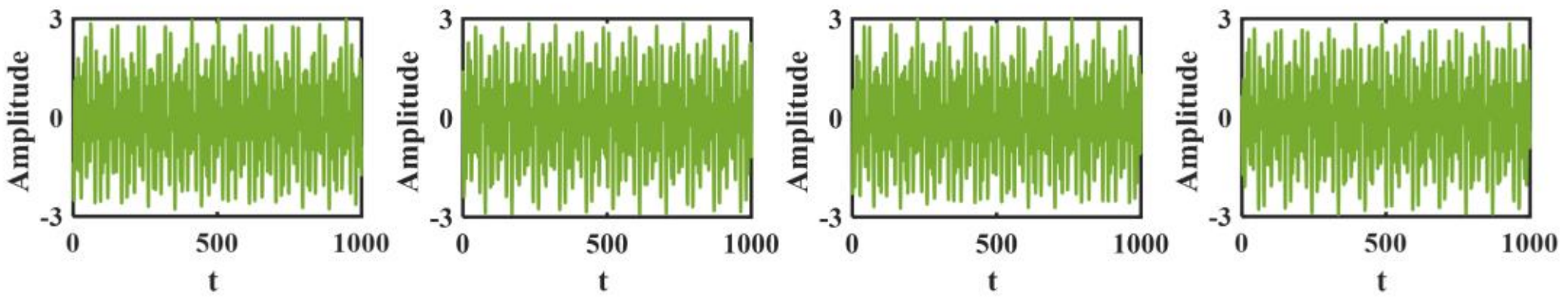

(a) (b) (c) (d)

Fig. 12 Partial time series in the multi-frequency dataset.

During the forward diffusion and training process, the batch size is set to 10, and the TSDM is trained over 200 epochs to realize denoising generation. The number of noise diffusion and denoising layers T is set to 3000. Based on the trained model, 40 target time series are generated, a partial of the results is shown in Fig. 13, and the corresponding frequency spectrum is shown in Fig. 14.

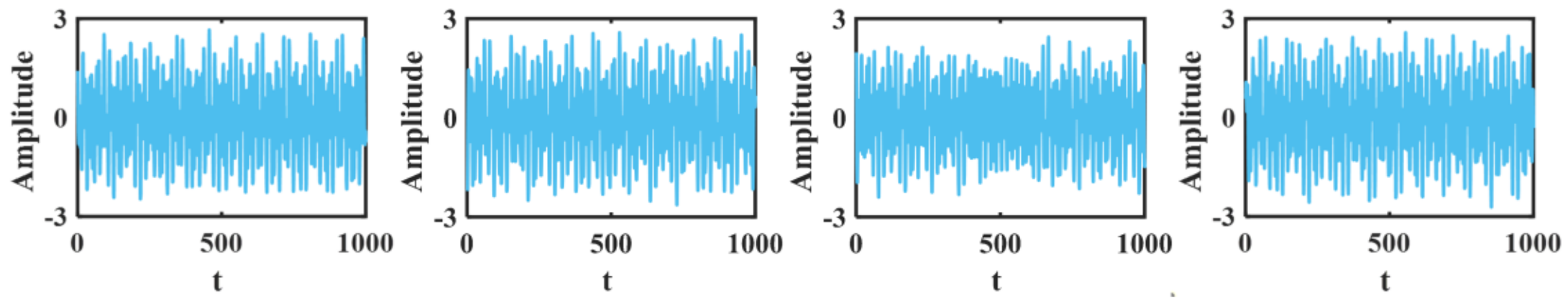

(a) (b) (c) (d)

Fig. 13 Generation results of multi-frequency dataset

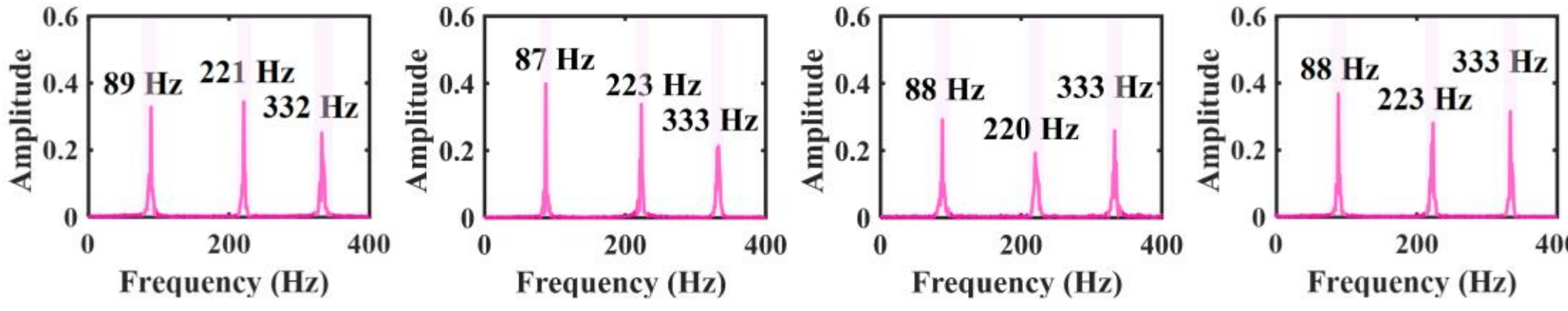

(a) (b) (c) (d)

Fig. 14 Corresponding frequency spectrum of generation results of multi-frequency dataset.

In the diffusion generation results of the multi-frequency dataset, it can be seen in Fig. 13 that the time series generated by diffusion show standard beat characteristics, which often appear in multi-frequency series, and the periodicities are consistent. Fig. 14 shows the corresponding frequency spectrum of generation results, it can be seen that the characteristic frequency of 88Hz, 222Hz and 333Hz are preserved after diffusion generation, which reflect the accuracy of generated results. Due to the high frequency, the frequency of some generated results has an error of no more than 2%, which is acceptable. In the spectrum of multi-frequency generation results, the randomness of generation results is more obvious than that of single-frequency generation results. The bandwidth and amplitude of the three characteristic frequencies are different between generation results. That also means the generated results of the TSDM have certain randomness while retaining the main characteristics. It also shows that TSDM can generate diversified target time series instead of simply copying training samples after multi-frequency time series training.

Summarize the frequency spectrums of 40 time series generated and draw the box plot as shown in Fig. 15. It can be seen from the box plot that the peak values of the frequency component of the generated time series are mainly around the characteristic frequency. The amplitude fluctuation of the peak is relatively large compared with other frequency positions, and it can be seen that the bandwidth of the average spectrum peak is significantly wider than that of a single sample. The uncertainty of the amplitude and bandwidth of the resulting spectrum peak reflects the creativity of the TSDM. In addition, it can be seen that in the multi-frequency generation results, the number of outliers is far more than that in the single-frequency generation results, which also shows that TSDM is more creative for the target time series with more features.

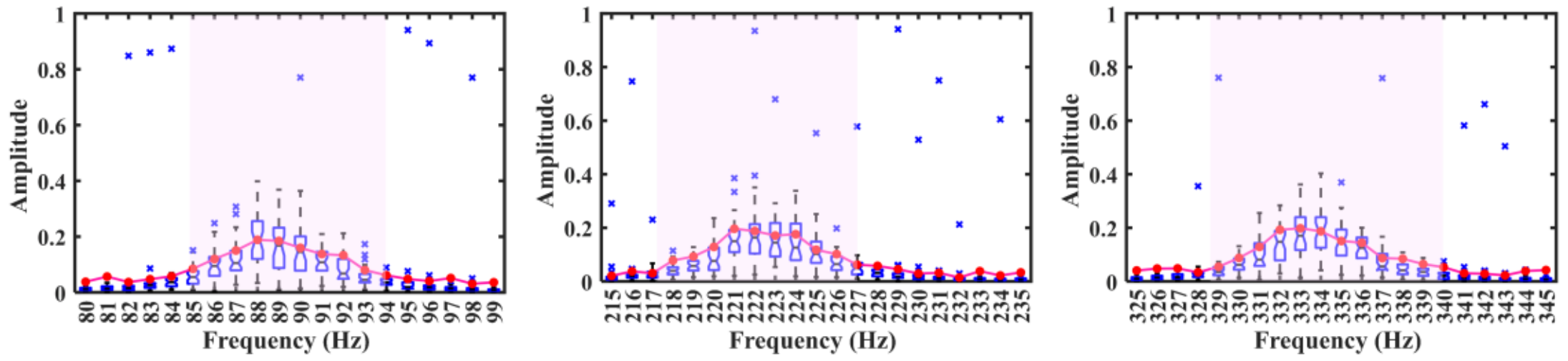

(a) (b) (c)

Fig. 15 Box plot of generation results of multi-frequency dataset of (a) 88Hz. (b) 222Hz. (c) 333Hz.

Taking the generated result in Fig. 13(a) as an example, draw the process of gradual denoising it from random noise to generating a single-frequency trigonometric function, as shown in Fig. 16. From the denoising generation process, it can be seen that with the increase of denoising times *t*, the random noise first gradually forms the contour of the target sequence. It can be seen from the corresponding frequency spectrum that with the increase of denoising times, the corresponding peak of the characteristic frequency of the target series gradually appears and increases. Since there are three frequency components in the time series, the beat phenomenon of the series cannot be clearly seen until *t*=3000.

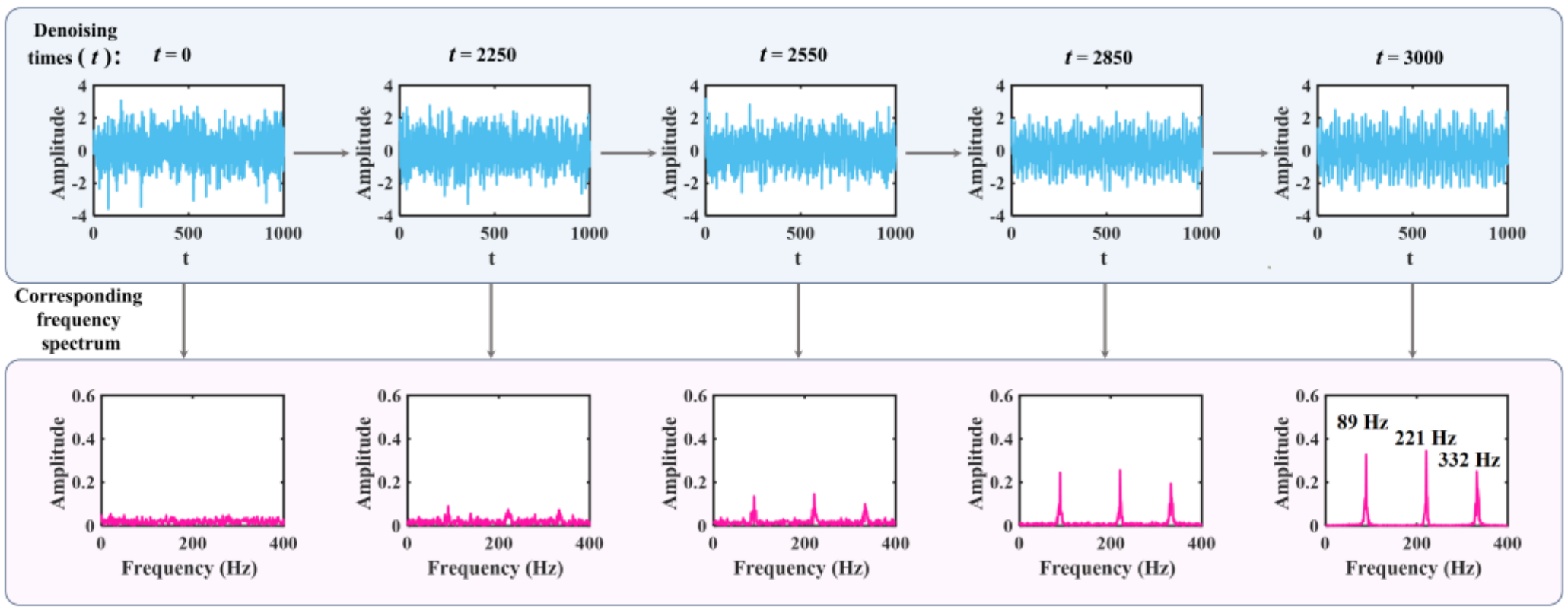

Fig. 16 Denoising generation process of TSDM for a multi-frequency data example

For the generation of multi-frequency time series, the TSDM proposed in this paper is also compared with the existing time series generation methods. The comparison focuses on the waveform coincidence degree, spectral consistency and VF value between the generated time series and the target time series. The results are shown in Fig. 17, Fig. 18 and Tab. 2.

In order to increase the difficulty of multi-frequency dataset, when constructing multi frequency time series, the phases of each frequency time series are different, which shows that in equation (15), the values of $b_{mn}$ change. Therefore, we cannot evaluate the generation quality by fitting the waveforms of the generated time series and the target time series. From the waveform results of the generated time series in Fig. 17, it can be seen that the generated results obtained by different methods seem to be similar to the original signal. But when comparing the spectrum of time series generated by different methods, the difference is obvious. It can be seen from Fig. 18 that for the spectrum of the time series generated by VQ-VAE, TimeGAN and Diffwave, the frequency band of the peak corresponding to the target frequency is wider and the frequency peak is smaller (compared with the target frequency peak of 0.5). In contrast, in the spectrum of the time series generated by TSDM, the frequency band of the frequency peak is narrower, the peak is larger and closer to the target value of 0.5. This fully shows that for the generation of multi-frequency time series, TSDM has higher accuracy than other existing methods. From the *VF* values of the four methods results in Tab. 2, it can also be seen that the *VF* values of the times series generated by the TSDM method are significantly lower than those of the other three methods, indicating that the accuracy of the TSDM generated results is higher and TSDM can preserve frequency characteristics better.

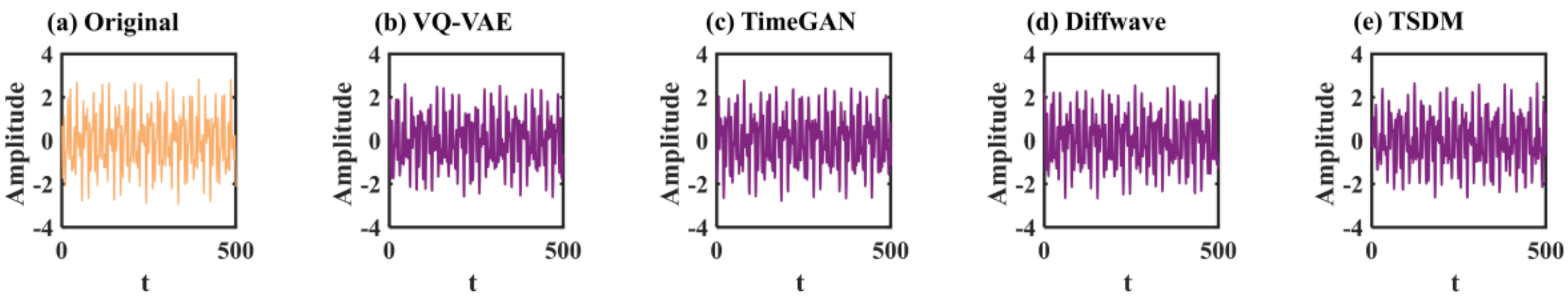

Fig. 17 Generated multi-frequency time series. (a) Original time series. (b) Generated by VQ-VAE. (c) Generated by TimeGAN. (d) Generated by Diffwave. (e) Generated by proposed TSDM.

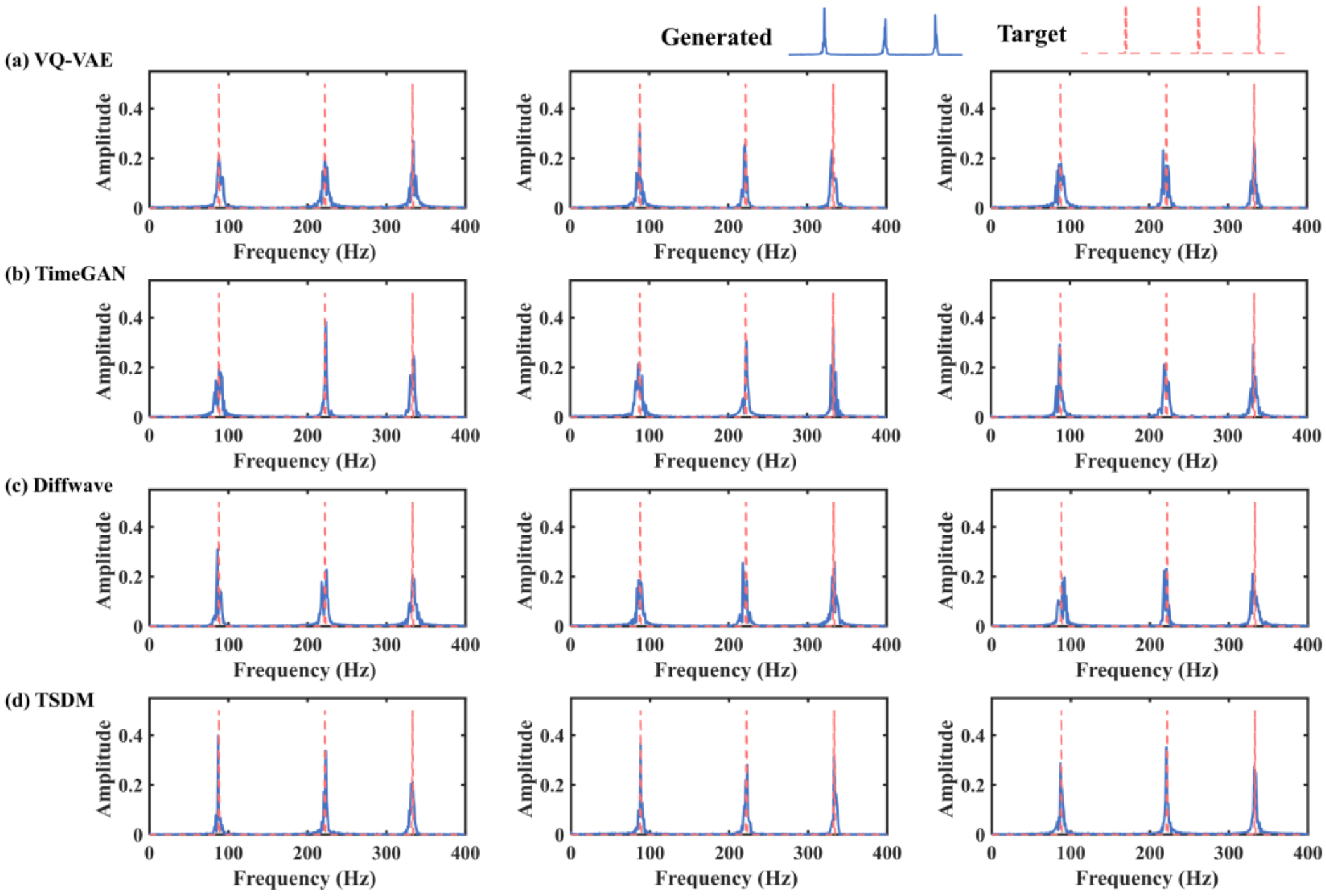

Fig. 18 Comparison of target signal spectrum and generated signal spectrum generated by (a) VQ-VAE. (b) TimeGAN. (c) Diffwave. (d) proposed TSDM.

Tab. 2 *VF* value of the time series generated by different methods.

| | VQ-VAE | TimeGAN | Diffwave | TSDM |
|---|---|---|---|---|
| Variance Frequency | 0.0690 | 0.0725 | 0.0661 | **0.0541** |

### 4.3 Bearing Fault Data

In Sec. 4.1 and 4.2, the excellent diffusion generation ability of TSDM for regular sequences is proved by artificially constructing single-frequency and multi-frequency time series datasets. In order to test the ability of TSDM to generate actual vibration signals, which also determines whether it can be applied in practice, this section selects a public bearing fault dataset to train TSDM and do diffusion generation. The selected XJTU bearing fault dataset includes outer ring fault, inner ring fault, cage fault and mixed fault. The dataset size is [200,2048], which contains 200 time series of each fault with a length of 2048. Partial time series in the XJTU dataset are shown in Fig. 19.

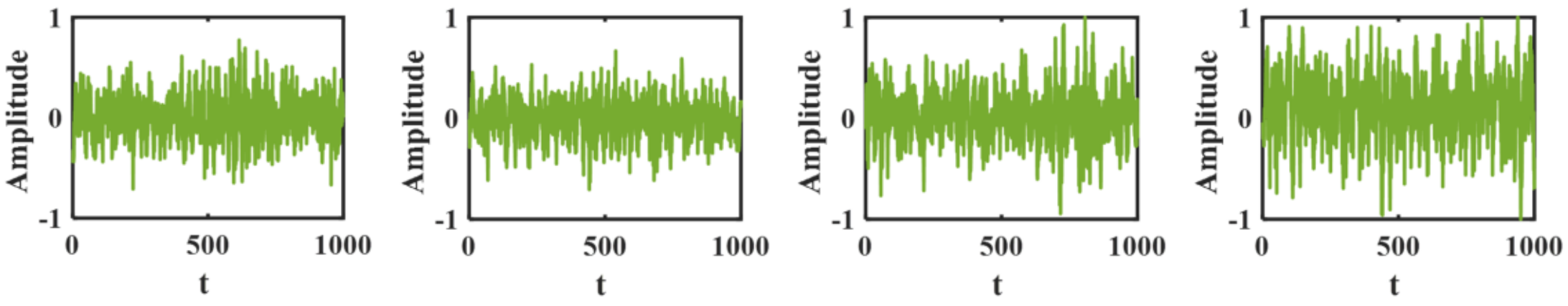

(a) (b) (c) (d)

Fig. 19 Partial time series in XJTU bearing fault dataset of (a) outer ring fault. (b) inner ring fault. (c) cage fault. (d) mixed fault.

During the forward diffusion and training process, the batch size is set to 10, and the TSDM is trained over 200 epochs to realize denoising generation. The number of noise diffusion and denoising layers T is set to 3000. Based on the trained model, 40 target time series of each fault are generated, and a partial of the results is shown in Fig. 20. Because the speed corresponding to the same fault data in the bearing fault dataset is different, it also leads to the fault characteristic frequency corresponding to the same fault may be different. To reflect the retention of the generated results on the fault characteristics, the frequency spectrum of the same fault in the dataset is summarized and averaged, and drawn in the same figure as the frequency spectrum of the generated results, as shown in Fig. 21.

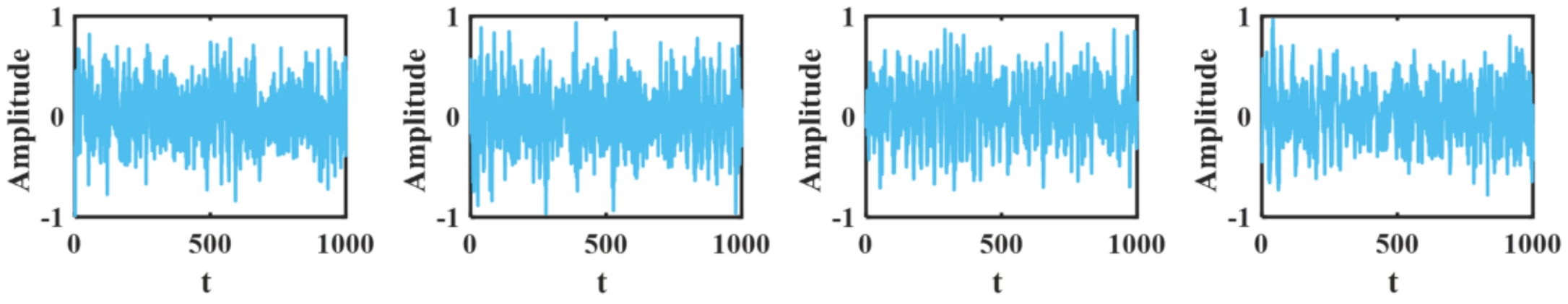

(a) (b) (c) (d)

Fig. 20 Generation results of XJTU bearing fault dataset of (a) outer ring fault. (b) inner ring fault. (c) cage fault. (d) mixed fault.

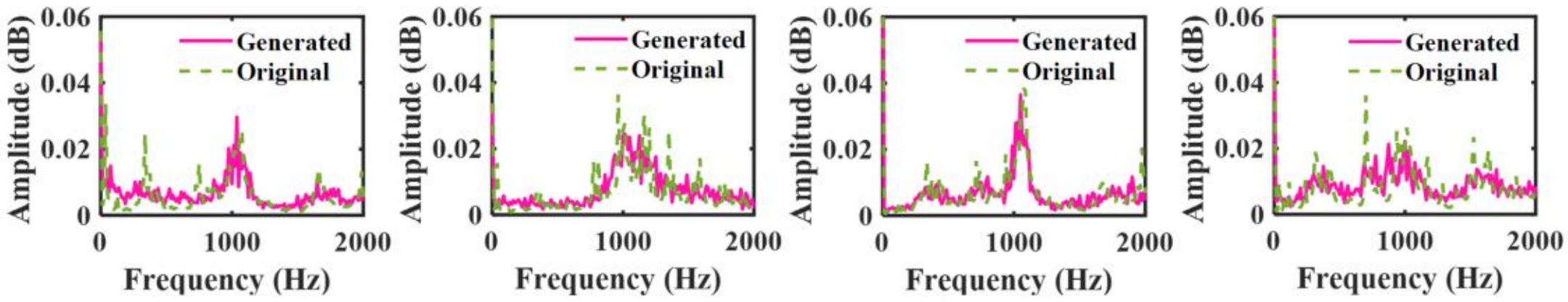

(a) (b) (c) (d)

Fig. 21 Corresponding frequency spectrum of generation results in XJTU dataset of (a) outer ring fault. (b) inner ring fault. (c) cage fault. (d) mixed fault.

In the diffusion generation results of XJTU bearing fault dataset, it can be seen in Fig. 19 that the time series generated by diffusion cannot be directly observed the fault characteristics, which is different from the results of single-frequency and multi-frequency time series datasets. This also means preserving and generating the features of bearing fault datasets is a more difficult task. In Fig. 21, the blue dotted line represents the average spectrum of 200 data in the training set, and the red solid line represents the spectrum of a single generated result. It can be seen that the spectral lines of the average frequency spectrum are relatively smooth, while the generated resulting spectral lines have more frequency components obviously. Overall, the frequency spectrum of the generated results is consistent with the average frequency spectrum trend of the training set, and the two have a high degree of coincidence. This shows that TSDM can generate bearing fault time series with similar characteristics to the training set. This also proves that TSDM can generate simple standard time series and measured data, which will significantly expand the application prospect of TSDM.

### 4.4 Ablation Study

To illustrate the superiority of the TSDM proposed in this paper, DDPM and other improved methods are used to compare with TSDM. The improvements of TSDM mainly focus on the network structure of U-net, which add ResBlock and AttnBlock to the up-sample module. The up-sampling module structure of DDPM, its separate combination with ResBlock and AttnBlock, and TSDM is shown in Fig. 22.

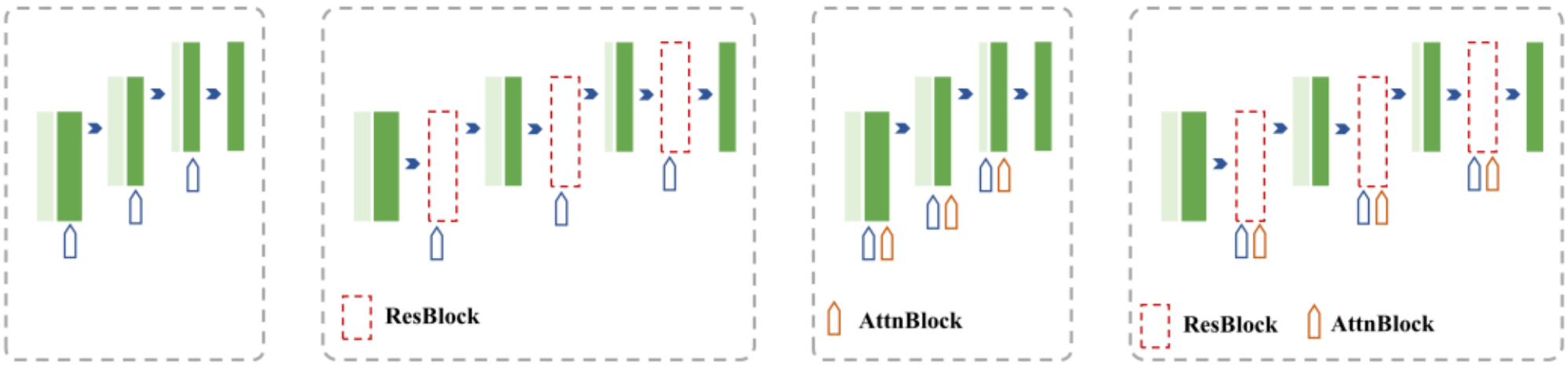

(a) (b) (c) (d)

Fig. 22 The network structure in the up-sampling module of (a) DDPM. (b) DDPM+ResBlock. (c) DDPM+AttnBlock. (d) TSDM.

To measure the complexity of the network, the FLOPs (floating point operations) and parameter size of (a) DDPM (b) DDPM+ResBlock (c) DDPM+AttnBlock and (d) TSDM are counted in Tab. 3, and the Variance Frequency is used to compare the effectiveness of different methods for generating vibration signals. The effect of the quantity of ResBlock and AttnBlock is also discussed in Tab. 3.

Tab. 3 Comparison of FLOPs, Parameter Size and Variance Frequency for different methods

| Method | Block | | FLOPs | Parameter Size | Variance Frequency |
|---|---|---|---|---|---|
| | ResBlock | AttnBlock | | | |
| baseline | 2 | [2] | 23.501G | 554.993K | 0.0299 |
| DDPM | 0 | 0 | 5.822G | 105.169K | 0.0638 |
| DDPM+ResBlock | 2 | 0 | 21.142G | 507.953K | 0.0366 |
| DDPM+AttnBlock | 0 | [2] | 6.305G | 114.577K | 0.0705 |
| TSDM | 2 | [0] | 22.191G | 513.393K | 0.0399 |
| | | [1] | 23.239G | 529.073K | 0.0345 |
| | | [2] | 23.501G | 554.993K | 0.0299 |
| | | [0, 1] | 24.288G | 534.513K | 0.0325 |
| | | [0, 2] | 24.550G | 560.433K | 0.0374 |
| | | [1, 2] | 25.598G | 576.113K | 0.0373 |
| | | [0, 1, 2] | 26.647G | 581.553K | 0.0362 |
| | 1 | [2] | 16.096G | 384.017K | 0.0460 |
| | 2 | | 23.501G | 554.993K | 0.0299 |
| | 3 | | 30.907G | 725.969K | 0.0297 |

As shown in Tab. 3, compared with the U-net in DDPM, the addition of ResBlock causes a large increase in FLOPs and Parameter Size, and the generation accuracy significantly increased, which is manifested by the decrease of Variance Frequency value. However, the addition of AttnBlock results in a slight increase in FLOPs and Parameter Size, but a slight decrease in generation accuracy, and AttnBlock seems to be a negative effect on the quality of vibration signal generation. But after TSDM introduces both AttnBlock and ResBlock at the same time, the generation quality has been significantly improved, especially when AttnBlock is added to the specific module, the optimal combination scheme of generation quality can be achieved. Although, when the number of ResBlock is 3, the value of Variance Frequency is the minimum, which means that the accuracy of the data generated by the model is the highest at this time. However, compared with the baseline we selected, under this set of parameters, the computing burden is significantly increased, but the accuracy is only slightly improved, so we did not choose this set of parameters as the baseline.

The variation curves of Variance Frequency values with denoising times of the four methods are shown in Fig. 23. It can be seen that for a total of 3000 times of denoising generation, there is a significant effect starting from the 1500th generation. According to the trend of the curve, ResBlock has a significant impact on the generation accuracy, and on this basis, the addition of AttnBlock further improves the generation quality.

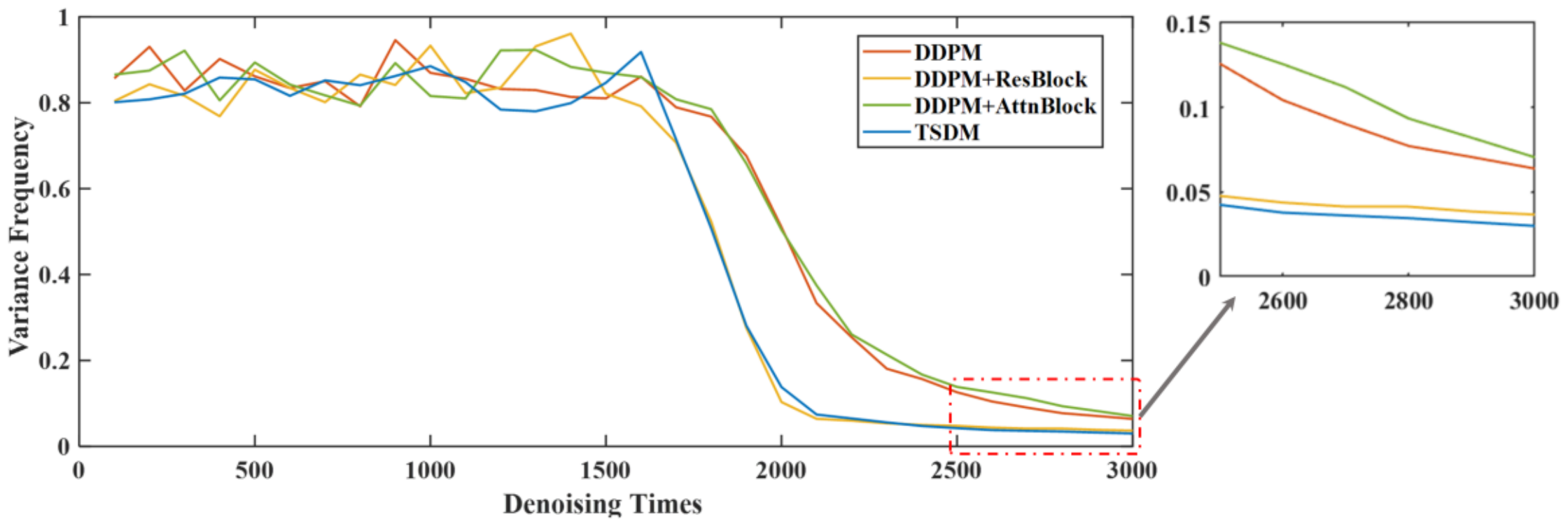

Fig. 23 The curves of Variance Frequency of different methods under the denoising times.

# 5. Practical Application in Small Sample Fault Diagnosis

In Sec. 4, the TSDM exhibits excellent generation ability for single-frequency time series, multi-frequency time series, and bearing fault data. In the actual fault diagnosis based on deep learning, the accuracy of diagnosis will be low due to the lack of training samples, called small sample fault diagnosis. Reasonable expansion of the small sample training set will effectively solve this problem. In this section, we define a case of small sample fault diagnosis and expand the small sample dataset through TSDM to improve fault diagnosis accuracy, as shown in Fig. 24.

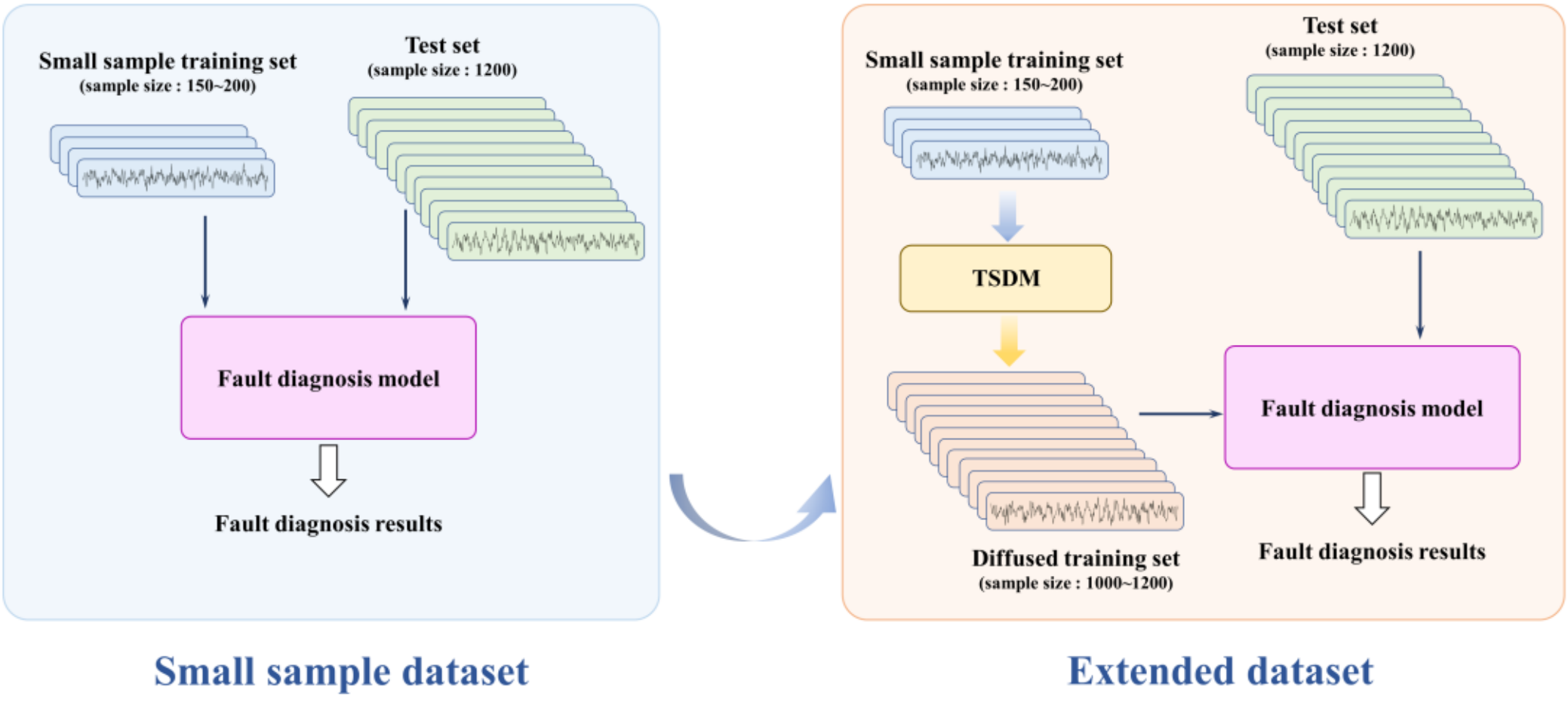

Fig. 24 Expansion of small sample dataset based on TSDM.

5.1 Small sample fault diagnosis under CWRU dataset[51]

CWRU bearing fault dataset is widely used in the field of bearing fault diagnosis. Researchers prefabricated three kinds of faults through Electro-Discharge Machining, including inner ring fault (IR), outer ring fault (OR) and rolling ball fault (RB), as well as a fault-free health state, in addition, a fault-free normal condition (NC) test was carried out. For the four working conditions of IR, OR, RB and NC, 50 samples of each working condition and a total of 200 samples are randomly selected as a small sample training set[50]; 300 samples for each working condition and a total of 1200 samples are selected as the test set. 400 samples of each working condition and a total of 1200 samples are generated based on small sample training set as the diffusion training set. The basic information of the small sample dataset used is shown in Tab. 4.

Tab. 4 Small sample dataset information of CWRU dataset.

| Datasets | Fault type | Number | Total | Length |
|---|---|---|---|---|
| Training set | IR | 50 | 200 | 2048 |
| | OR | 50 | | |
| | RB | 50 | | |
| | NC | 50 | | |
| Diffusion training set | IR | 250 | 1000 | |
| | OR | 250 | | |
| | RB | 250 | | |
| | NC | 250 | | |
| Test set | IR | 300 | 1200 | |
| | OR | 300 | | |
| | RB | 300 | | |
| | NC | 300 | | |

In this study, three machine learning methods, CNN[48], RNNLSTM[49] and TST[50], are selected to compare the fault diagnosis results before and after the diffusion of small sample dataset. The detailed structures of CNN, RNNLSTM and TST are shown in Tab. 5.

Tab. 5 Detailed structures of CNN, RNNLSTM and TST.

| **Model** | **Structure Parameters** | | | | | | | |
|---|---|---|---|---|---|---|---|---|
| CNN | $\begin{bmatrix}\text{Conv1D(1, 25, 256)}\\\text{BatchNorm(25)}\\\text{ReLU}\\\text{Maxpool1D(2, 2)}\end{bmatrix}\rightarrow\begin{bmatrix}\text{Conv1D(25, 50, 3)}\\\text{BatchNorm(50)}\\\text{ReLU}\\\text{Maxpool1D(2, 2)}\end{bmatrix}\rightarrow\begin{bmatrix}\text{Linear(22350, 1024, ReLU)}\\\text{Linear(1024, 128, ReLU)}\\\text{Linear(128, 10)}\end{bmatrix}$ | | | | | | | |
| RNN-LSTM | $\text{Conv1D(1, 128, 3)}\rightarrow\begin{bmatrix}\text{LSTM(45, 64,tanh)}\\\text{Dropout(0.1)}\end{bmatrix}\rightarrow\begin{bmatrix}\text{Linear(64, 128, GeLU)}\\\text{Dropout(0.1)}\\\text{Linear(128, 10, ReLU)}\end{bmatrix}$ | | | | | | | |
| TST | $\boldsymbol{N_s}$ | $\boldsymbol{L/N_s}$ | $\boldsymbol{dim}$ | $\boldsymbol{dim_{MLP}}$ | $\boldsymbol{d_k}$ | $\boldsymbol{h}$ | $\boldsymbol{depth}$ | **Pos encoding** |
| | 256 | 8 | 128 | 256 | 64 | 8 | 4 | 1 D |

The batch size is set to 10, and the machine learning methods are trained over 100

epochs and repeated 50 times respectively. Before and after using the diffusion training set, the accuracy and loss function of the training and test set in the training process are shown in Fig. 25 to Fig. 27.

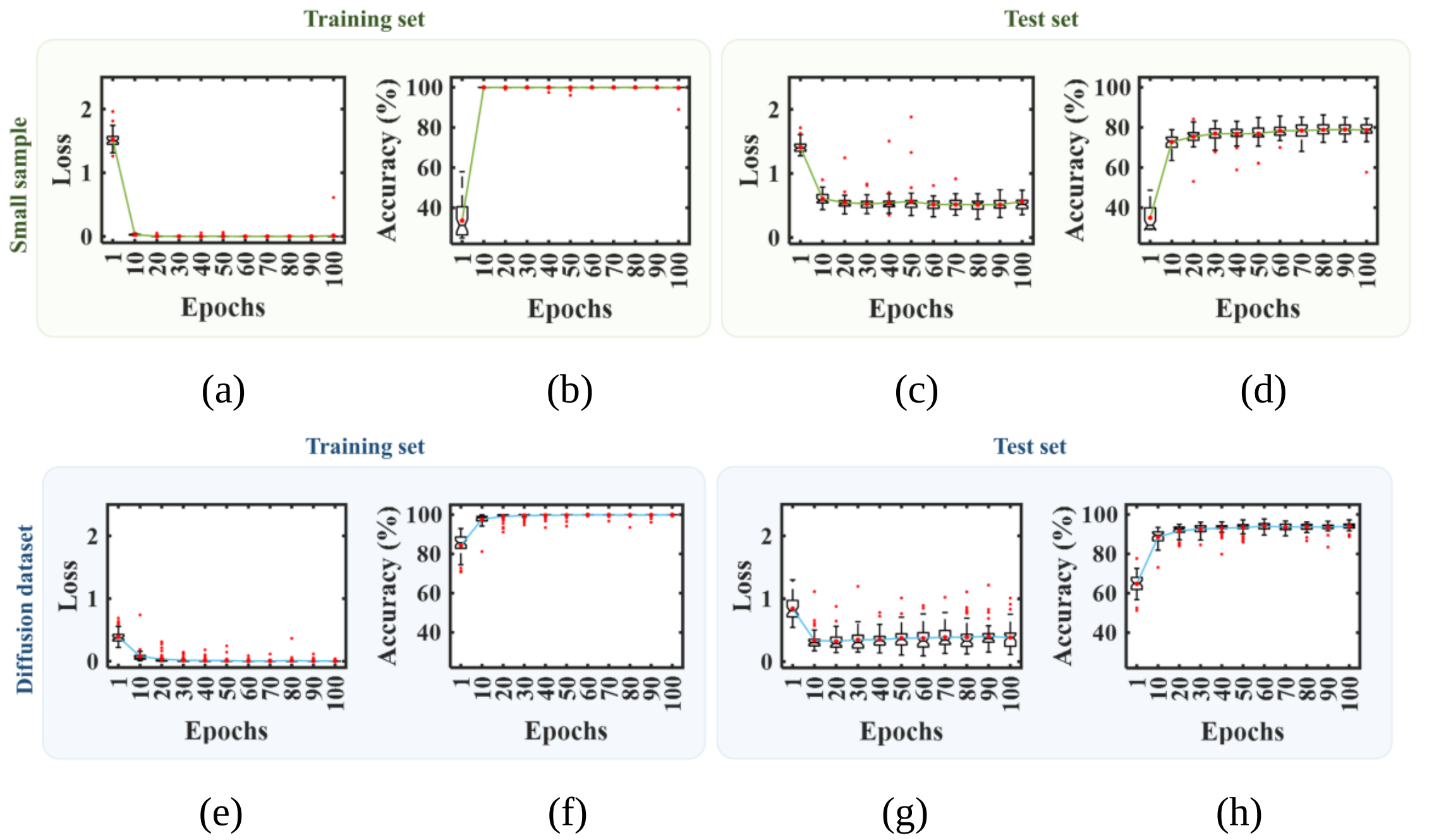

Fig. 25 Box plot of CNN training process under CWRU dataset. (a) Loss function of training set under small sample dataset. (b) Accuracy of training set under small sample dataset. (c) Loss function of test set under small sample dataset. (d) Accuracy of test set under small sample dataset. (e) Loss function of training set under diffusion training set. (f) Accuracy of training set under diffusion training set. (g) Loss function of test set under diffusion training set. (h) Accuracy of test set under diffusion training set.

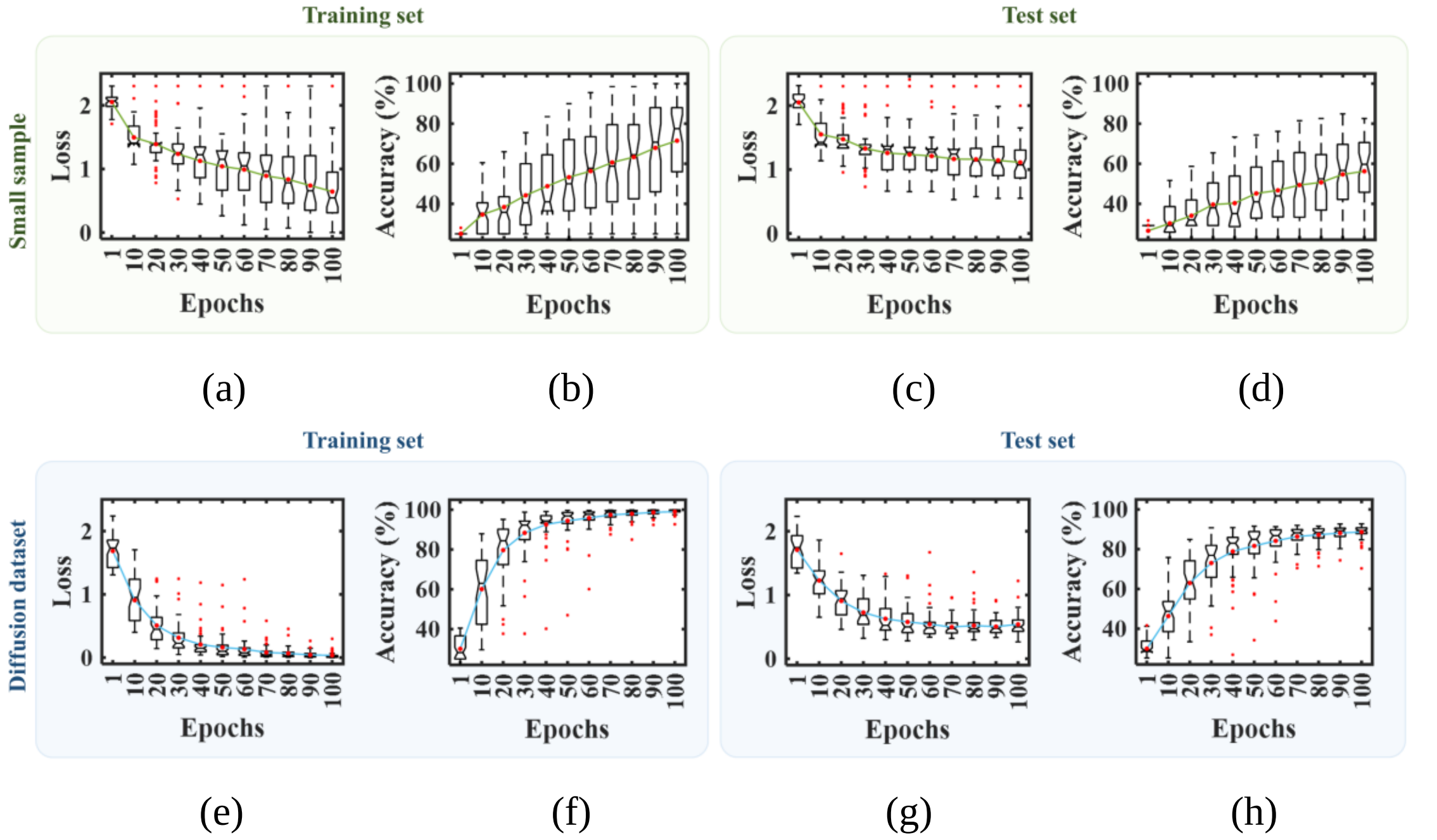

Fig. 26 Box plot of RNNLSTM training process under CWRU dataset. (a) Loss function of training set under small sample dataset. (b) Accuracy of training set under small sample

dataset. (c) Loss function of test set under small sample dataset. (d) Accuracy of test set under small sample dataset. (e) Loss function of training set under diffusion training set. (f) Accuracy of training set under diffusion training set. (g) Loss function of test set under diffusion training set. (h) Accuracy of test set under diffusion training set.

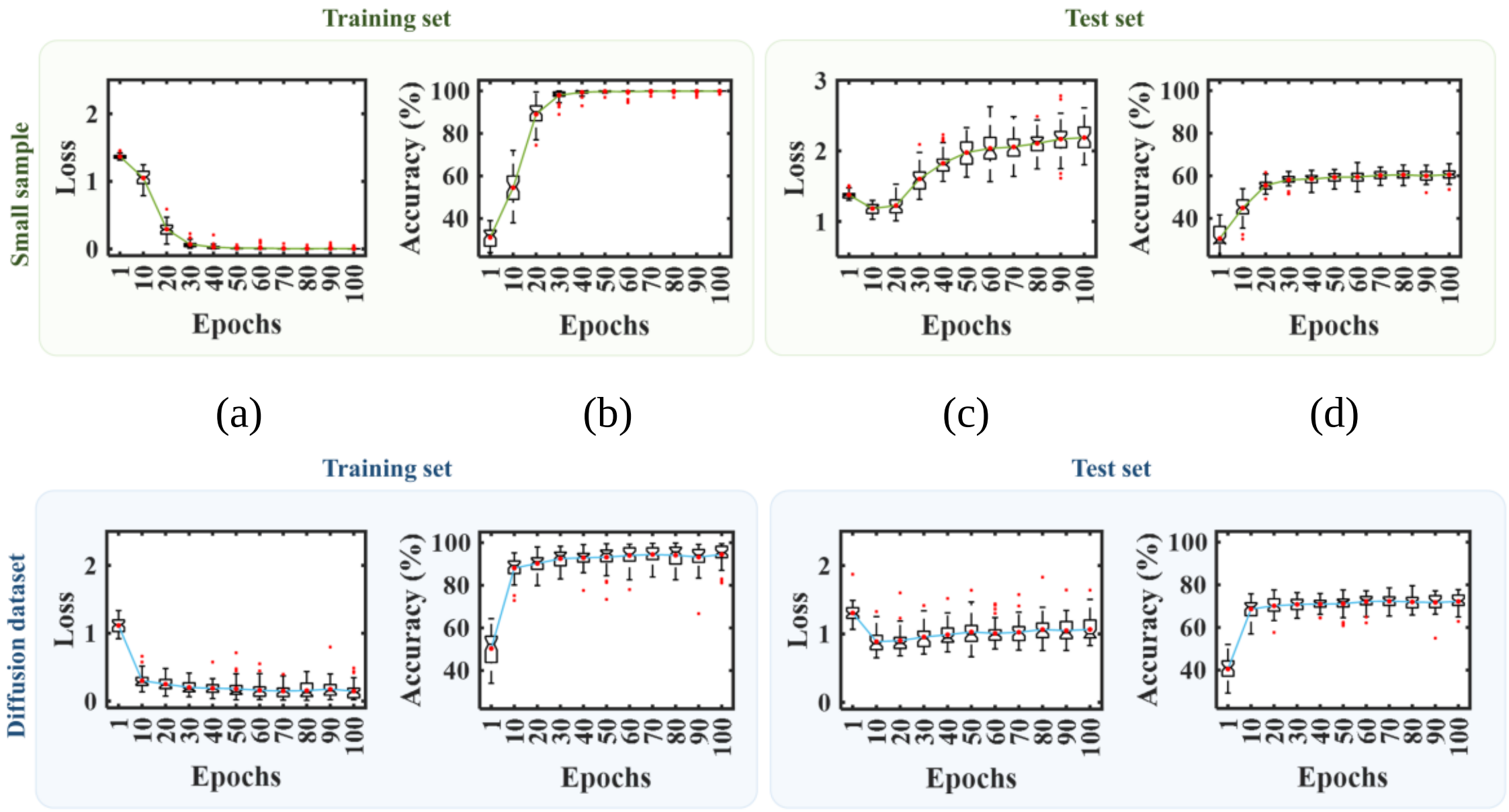

(a) (b) (c) (d)

(e) (f) (g) (h)

Fig. 27 Box plot of TST training process under CWRU dataset. (a) Loss function of training set under small sample dataset. (b) Accuracy of training set under small sample dataset. (c) Loss function of test set under small sample dataset. (d) Accuracy of test set under small sample dataset. (e) Loss function of training set under diffusion training set. (f) Accuracy of training set under diffusion training set. (g) Loss function of test set under diffusion training set. (h) Accuracy of test set under diffusion training set.

From the training process of CNN in Fig. 25, it can be seen that the diagnosis accuracy and loss function of the training set based on the diffusion training set reach the equilibrium position faster compared with small sample dataset. The diagnosis accuracy of the test set increases and the loss function decreases significantly. Although the diffusion of the dataset causes the increase of outliers in the training process based on the diffusion training set, the overall diagnosis results of CNN are positively improved.

From the training process of RNNLSTM in Fig. 26, it can be seen that the training results based on small sample dataset are seriously discrete, which is reflected in the box plot that the box is too long, especially in Fig. 26(a), (b) and (d). At the same time, the small sample training set causes RNNLSTM not to converge before epoch=100. These problems have been improved after using the diffusion dataset. As can be seen from Fig. 26(e), (f), (g) and (h), the loss function in the training process is reduced and the diagnosis accuracy is increased significantly. At the same time, the training process shows a good convergence trend.

From the training process of TST in Fig. 27, it can be seen that the results of TST mainly have the problems of increasing the test set loss function and low diagnosis accuracy in training based on small sample datasets, as shown in Fig. 27(c) and (d). After using the diffusion training set, these two problems have been improved, with the loss function gradually decreasing in Fig. 27(g) and the accuracy slightly improving in Fig. 27(h). In addition, the loss function and accuracy of the training set after using the diffusion training set converge faster than small sample dataset.

The accuracy of the test set at epoch=100 is summarized to reflect the contribution of TSDM and other methods, the box plot is shown in Fig. 28, and the summary table is shown in Tab. 6. It can be seen from Fig. 28 that the application of TSDM to expand the training set can effectively improve the accuracy of small sample fault diagnosis. The other three methods can also improve the accuracy, but the effect is obviously not as good as TSDM. The improvement of proposed TSDM ranges are 15.368%, 32.380% and 11.635% over small sample dataset respectively. The specific diagnostic accuracy results are shown in Tab. 6.

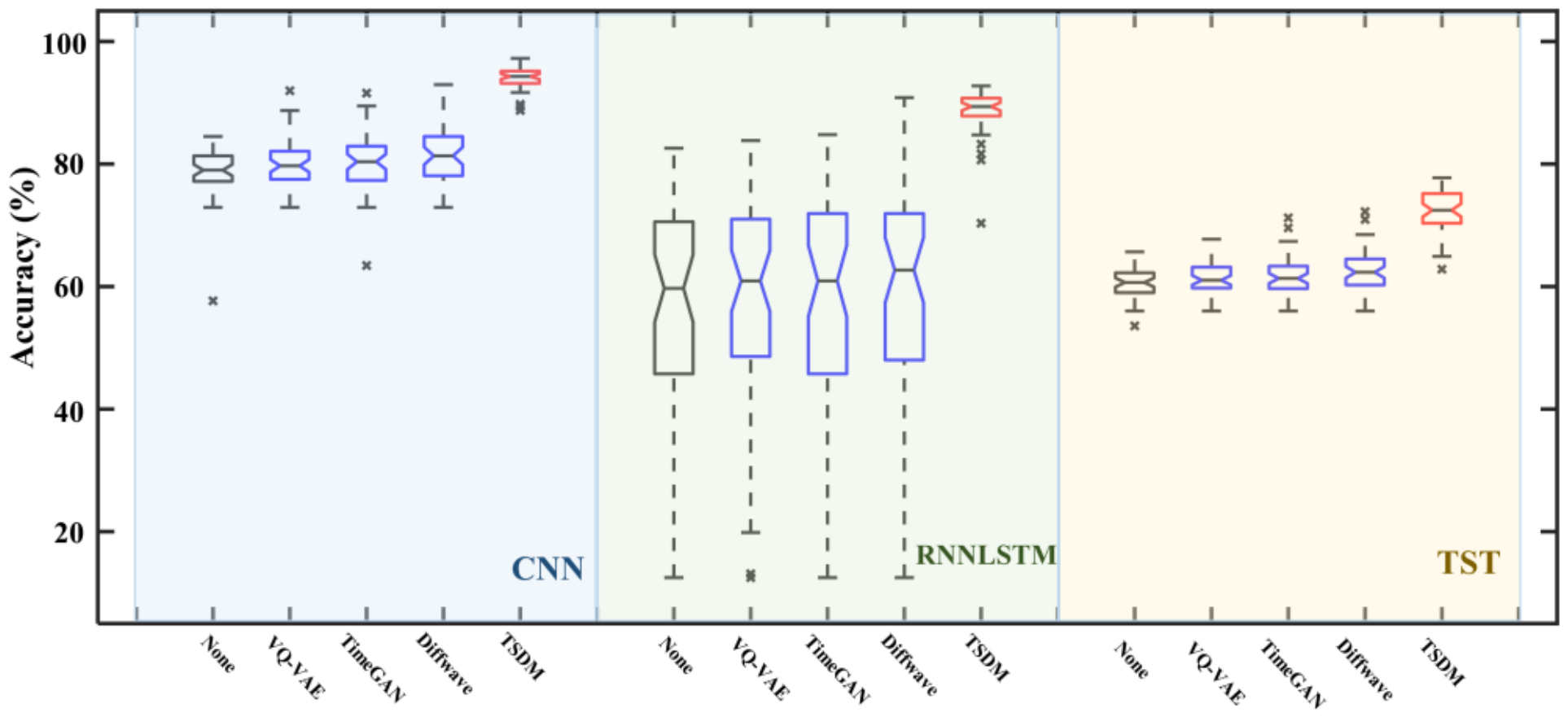

Fig. 28 Box plot of TSDM and other methods effect on test set accuracy under CWRU dataset at epoch=100.

Tab. 6 Accuracy of test set improved by TSDM and other methods under CWRU dataset at epoch=100.

| Method | Accuracy of test set | | | | | |
|---|---|---|---|---|---|---|
| | None | VQ-VAE | TimeGAN | Diffwave | TSDM | TSDM improved |
| CNN | 78.665% | 79.708% | 80.375% | 81.330% | 94.033% | 15.368% |
| RNNLSTM | 56.262% | 60.917% | 60.967% | 62.683% | 88.642% | 32.380% |
| TST | 60.557% | 61.038% | 61.333% | 62.330% | 72.192% | 11.635% |

5.2 Small sample fault diagnosis under XJTU dataset

XJTU bearing fault dataset is also widely used in the field of bearing fault diagnosis. It is a bearing fatigue fault dataset that contains data from 15 bearings operating until fatigue fault. The dataset includes four working conditions: inner ring fault (IR), outer ring fault (OR), cage fault (C), and mixed fault of inner ring, ball, outer ring and cage (IBOC). For the four working conditions of IR, OR, C and IBOC, 50 samples of each working condition and a total of 200 samples are randomly selected as small sample training set[50]; 300 samples for each working condition and a total of 1200 samples are selected as the test set. 250 samples of each working condition and a total of 1000 samples are generated based on small sample training set as the diffusion training set. The basic information of the small sample dataset used is shown in Tab. 7.

Tab. 7 Small sample dataset information of XJTU dataset.

| Datasets | Fault type | Number | Total | Length |
|---|---|---|---|---|
| Training set | IR | 50 | 200 | 2048 |
| | OR | 50 | | |
| | C | 50 | | |
| | IBOC | 50 | | |
| Diffusion training set | IR | 250 | 1000 | |
| | OR | 250 | | |
| | C | 250 | | |
| | IBOC | 250 | | |
| Test set | IR | 300 | 1200 | |
| | OR | 300 | | |
| | C | 300 | | |
| | IBOC | 300 | | |

CNN, RNNLSTM and TST are selected to compare the fault diagnosis results before and after using the diffusion training set. The batch size is set to 10, and the machine learning methods are trained over 100 epochs and repeated 50 times respectively. Before and after using the diffusion training set, the accuracy and loss function of the training and test set in the training process are shown in Fig. 29 to Fig. 31.

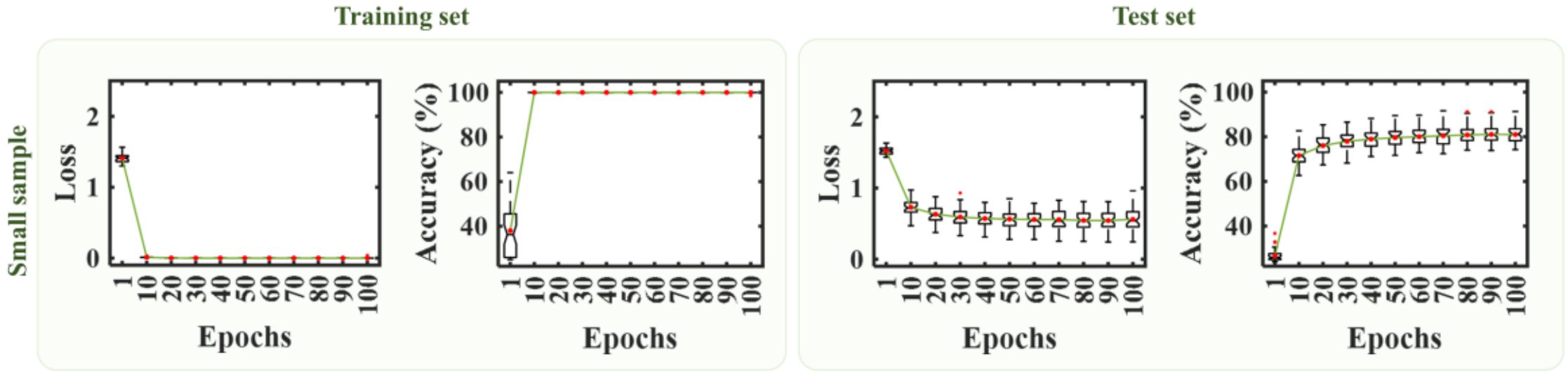

(a) (b) (c) (d)

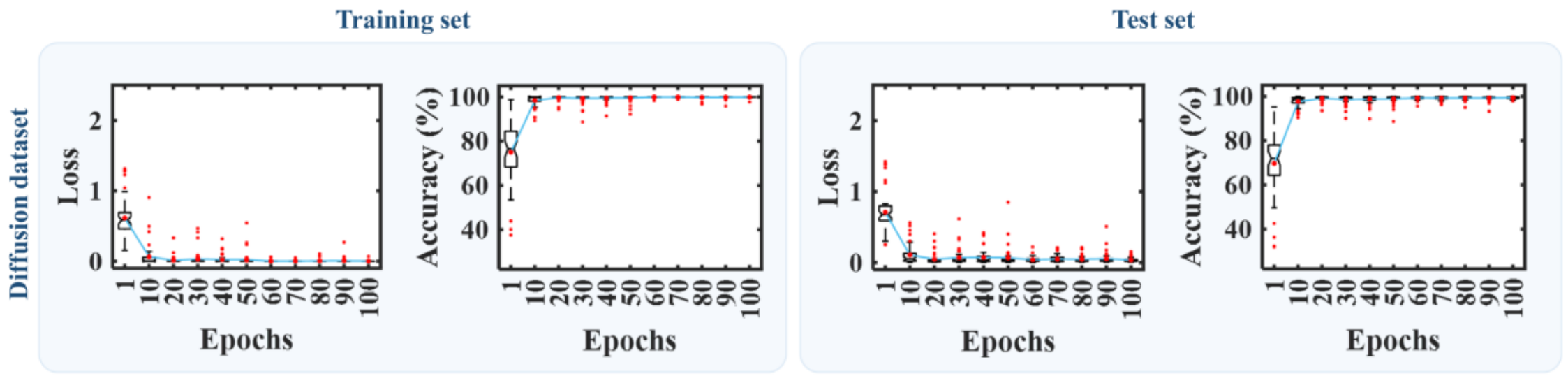

(e) (f) (g) (h)

Fig. 29 Box plot of CNN training process under XJTU dataset. (a) Loss function of training set under small sample dataset. (b) Accuracy of training set under small sample dataset. (c) Loss function of test set under small sample dataset. (d) Accuracy of test set under small sample dataset. (e) Loss function of training set under diffusion training set. (f) Accuracy of training set under diffusion training set. (g) Loss function of test set under diffusion training set. (h) Accuracy of test set under diffusion training set.

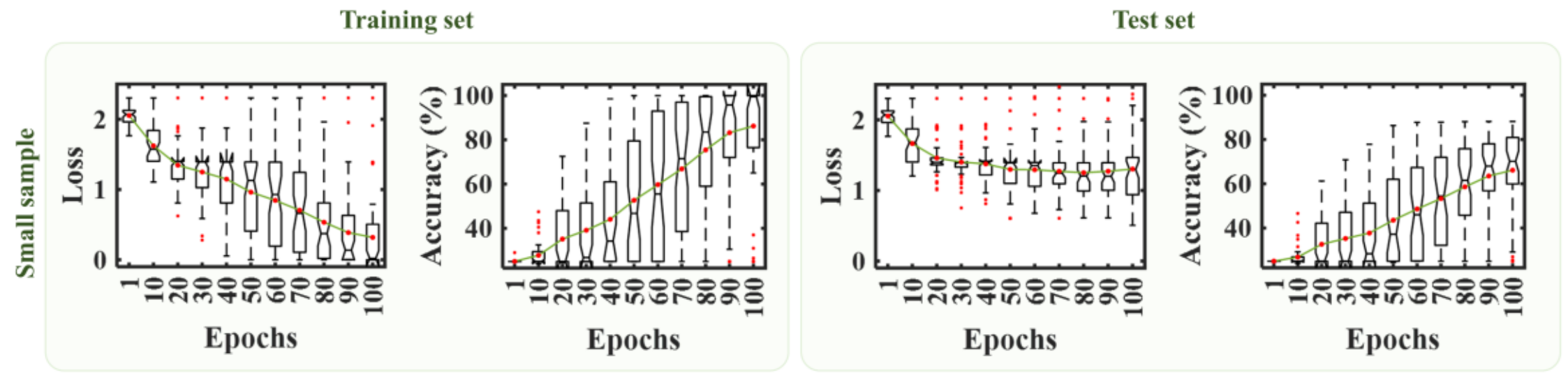

(a) (b) (c) (d)

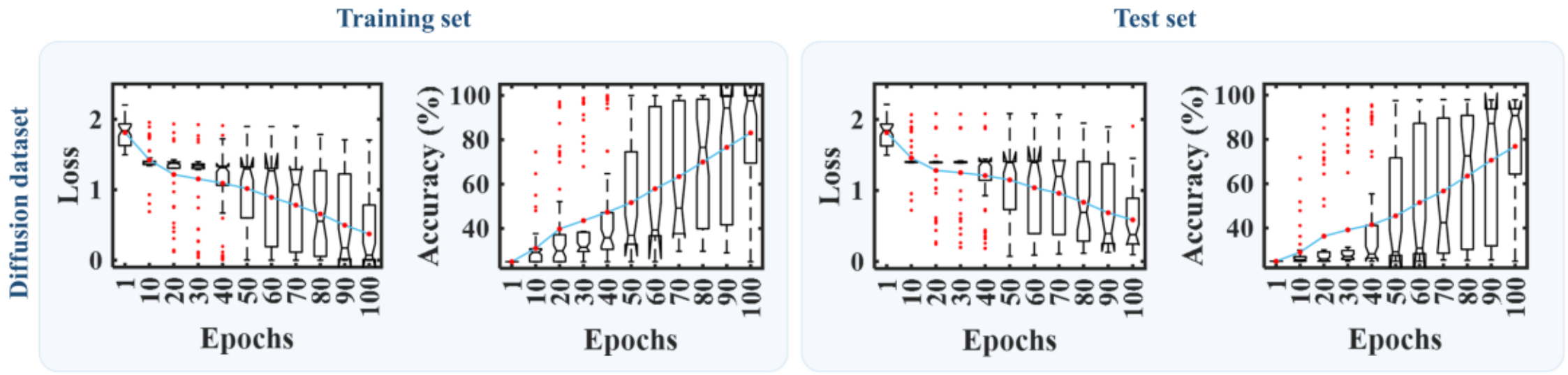

(e) (f) (g) (h)

Fig. 30 Box plot of RNNLSTM training process under XJTU dataset. (a) Loss function of training set under small sample dataset. (b) Accuracy of training set under small sample dataset. (c) Loss function of test set under small sample dataset. (d) Accuracy of test set under small sample dataset. (e) Loss function of training set under diffusion training set. (f) Accuracy of training set under diffusion training set. (g) Loss function of test set under diffusion training set. (h) Accuracy of test set under diffusion training set.

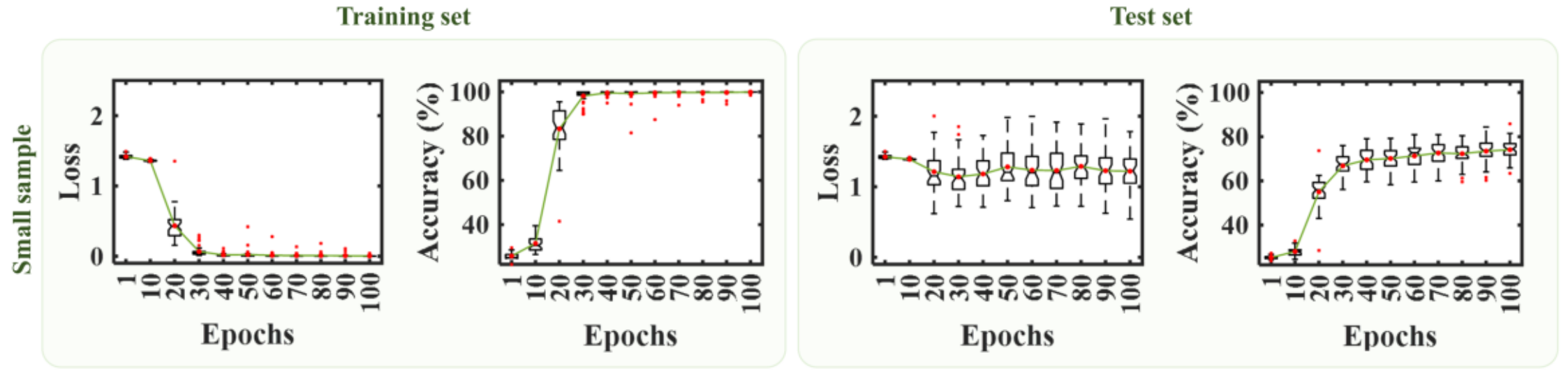

(a) (b) (c) (d)

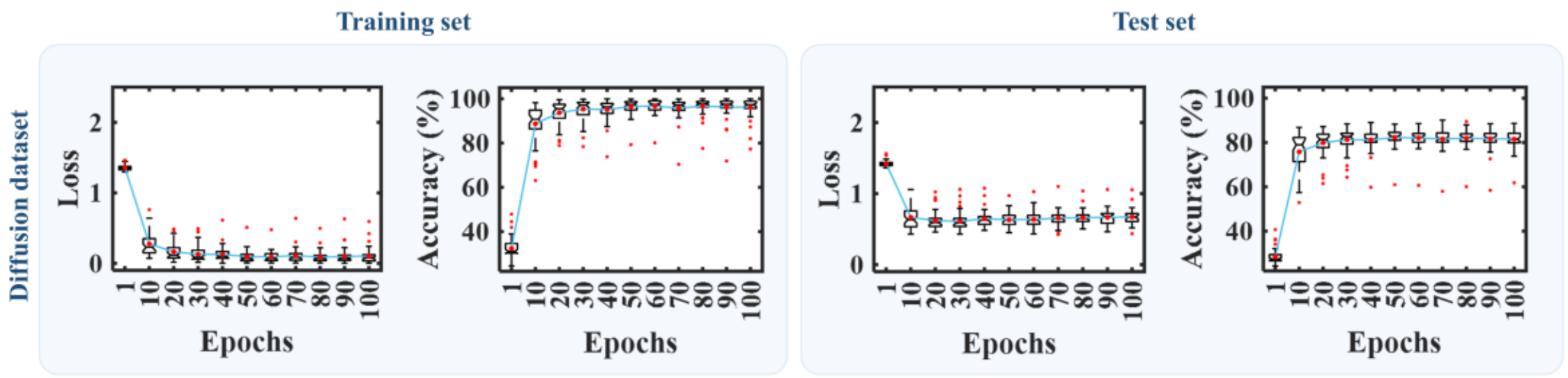

(e) (f) (g) (h)

Fig. 31 Box plot of TST training process under XJTU dataset. (a) Loss function of training set under small sample dataset. (b) Accuracy of training set under small sample dataset. (c) Loss function of test set under small sample dataset. (d) Accuracy of test set under small sample dataset. (e) Loss function of training set under diffusion training set. (f) Accuracy of training set under diffusion training set. (g) Loss function of test set under diffusion training set. (h) Accuracy of test set under diffusion training set.

From the training process of CNN in Fig. 29, the situation is similar to that under CWRU dataset, it can be seen that the diagnosis accuracy and loss function of the training set based on the diffusion training set reach the equilibrium position faster, compared with small sample dataset. The diagnosis accuracy of the test set increases and the loss function decreases significantly. Although the diffusion of the dataset causes the increase of outliers in the training process based on the diffusion training set, the overall diagnosis results of CNN are significantly improved.

From the training process of RNNLSTM in Fig. 30, it can be seen that the diagnosis results of RNNLSTM under small sample dataset of XJTU are abysmal, which is mainly reflected in the fact that the accuracy and loss function shown in Fig. 30(a), (b) and (d) do not converge before epoch=100, and the statistical results shown in Fig. 30(c) have too many outliers. The situation improved slightly after training with diffusion datasets, such as the loss function decreased, and the accuracy increased. However, it does not obviously improve the problem of poor convergence of loss function and accuracy. Nevertheless, applying the diffusion dataset has improved the accuracy of test sets.

From the training process of TST in Fig. 35, it can be seen that the loss function and accuracy of TST converge slowly before epoch=10 where the resulting curve changes gently, as shown in Fig. 31(a), (b) and (d). The loss function of the test set is hard to decline because of the overfitting phenomenon caused by the training of small sample dataset, as shown in Fig. 31(c). After using the diffusion training set, these problems have been improved, and the loss function and accuracy of the training set converge faster than small sample dataset. In addition, the accuracy of the test set has also been significantly improved.

The accuracy of the test set at epoch=100 is summarized to reflect the contribution of TSDM and other methods, the box plot is shown in Fig. 28, and the summary table is shown in Tab. 6. It can be seen from Fig. 28 that the application of TSDM to expand the training set can effectively improve the accuracy of small sample fault diagnosis. The other three methods can also improve the accuracy, but the effect is obviously not as good as TSDM. The improvement of proposed TSDM ranges are 15.368%, 32.380% and 11.635% over small sample dataset respectively. The specific diagnostic accuracy results are shown in Tab. 8.

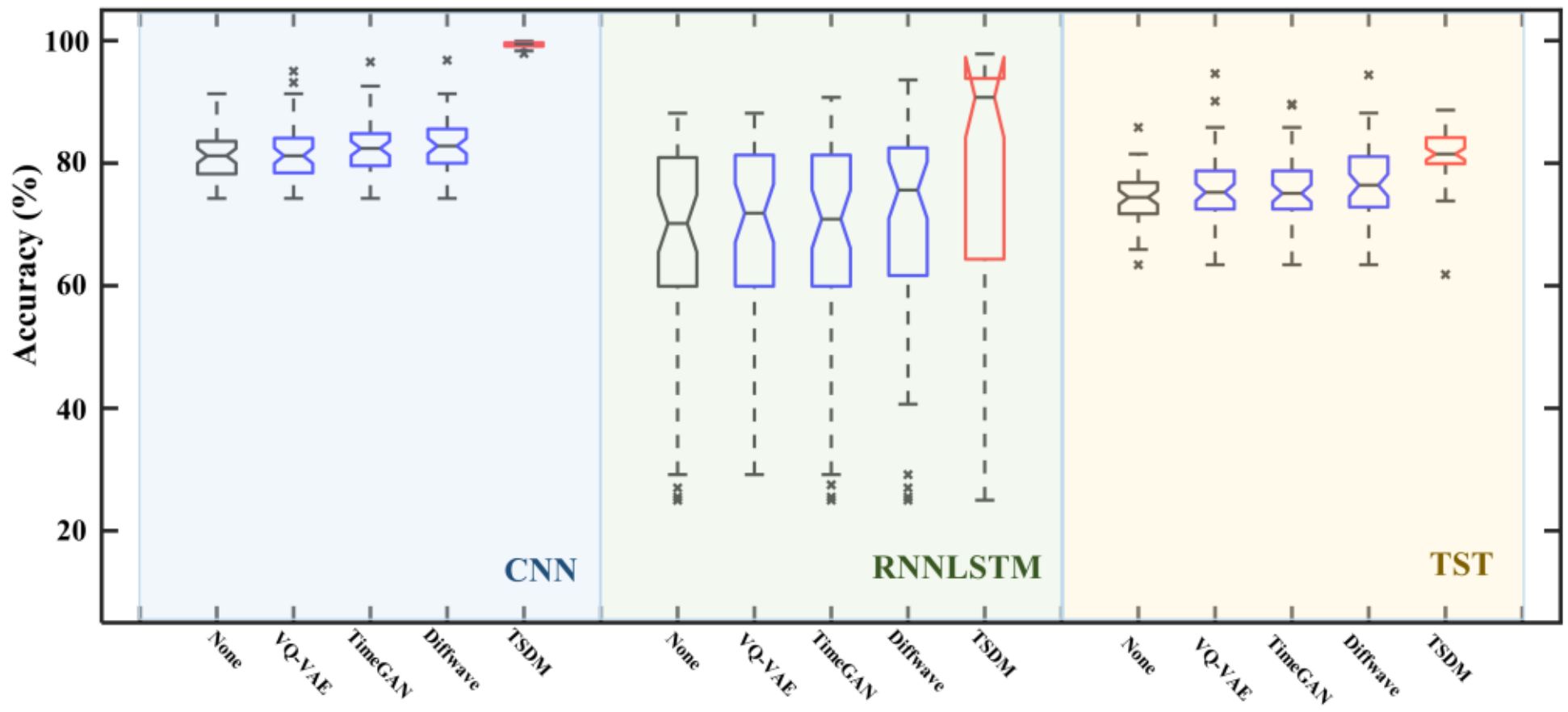

Fig. 32 Box plot of TSDM and other methods effect on test set accuracy under XJTU dataset at epoch=100.

Tab. 8 Accuracy of test set improved by TSDM and other methods under XJTU dataset at epoch=100.

| Method | Accuracy of test set | | | | | |
|---|---|---|---|---|---|---|
| | None | VQ-VAE | TimeGAN | Diffwave | TSDM | TSDM improved |
| CNN | 81.008% | 81.212% | 82.417% | 82.811% | 99.363% | 18.355% |
| RNNLSTM | 66.148% | 71.846% | 70.875% | 75.608% | 76.880% | 10.732% |
| TST | 74.113% | 75.250% | 75.083% | 76.417% | 81.562% | 7.449% |

## 5.3 Small sample fault diagnosis under HIT dataset[53]

HIT dataset is an inter-shaft bearing fault dataset based on an aero-engine system, which is obtained by the researchers of Harbin Institute of Technology through aero-engine test[54] and data processing[55]. The HIT dataset includes three working conditions: inner ring fault (IR), outer ring fault (OR) and normal condition (NC). For the three working conditions of IR, OR and NC, 50 samples of each working condition and a total of 150 samples are randomly selected as small sample training set[50]; 400 samples for each working condition and a total of 1200 samples are selected as the test set. 400 samples of each working condition and a total of 1200 samples are generated based on small sample training set as the diffusion training set. The basic information of the small sample dataset used is shown in Tab. 9.

Tab. 9 Small sample dataset information of HIT dataset.

| Datasets | Fault type | Number | Total | Length |
|---|---|---|---|---|
| | IR | 50 | | |
| Training set | OR | 50 | 150 | |
| | NC | 50 | | |
| | IR | 400 | | |
| Diffusion training set | OR | 400 | 1200 | 2048 |
| | NC | 400 | | |
| | IR | 400 | | |
| Test set | OR | 400 | 1200 | |
| | NC | 400 | | |

CNN, RNNLSTM and TST are selected to compare the fault diagnosis results before and after using the diffusion training set. The batch size is set to 10, and the machine learning methods are trained over 100 epochs and repeated 50 times respectively. Before and after using the diffusion training set, the accuracy and loss function of the training and test set in the training process are shown in Fig. 33 to Fig. 35.

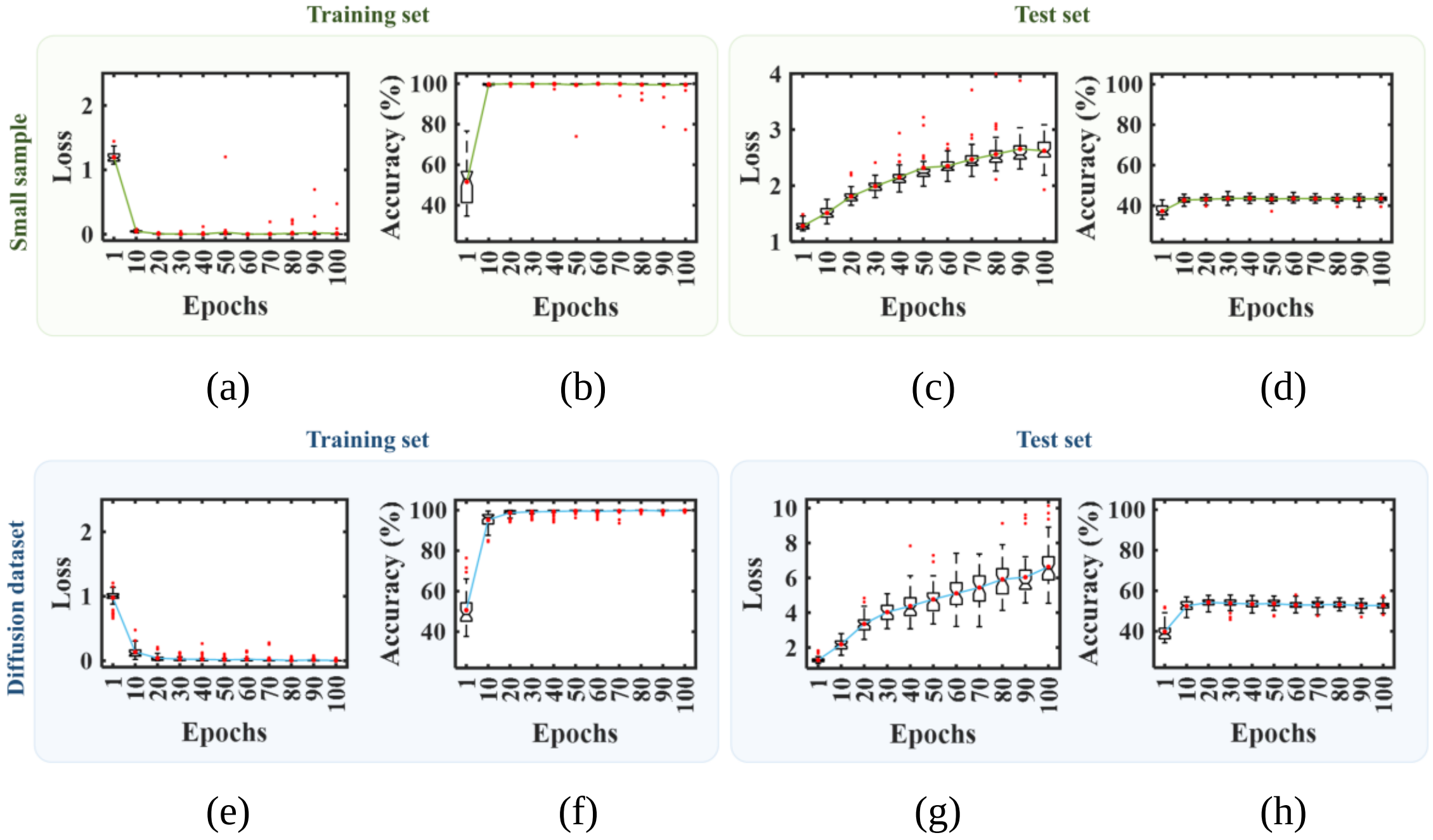

Fig. 33 Box plot of CNN training process under HIT dataset. (a) Loss function of training set under small sample dataset. (b) Accuracy of training set under small sample dataset. (c) Loss function of test set under small sample dataset. (d) Accuracy of test set under small sample dataset. (e) Loss function of training set under diffusion training set. (f) Accuracy of training set under diffusion training set. (g) Loss function of test set under diffusion training set. (h) Accuracy of test set under diffusion training set.

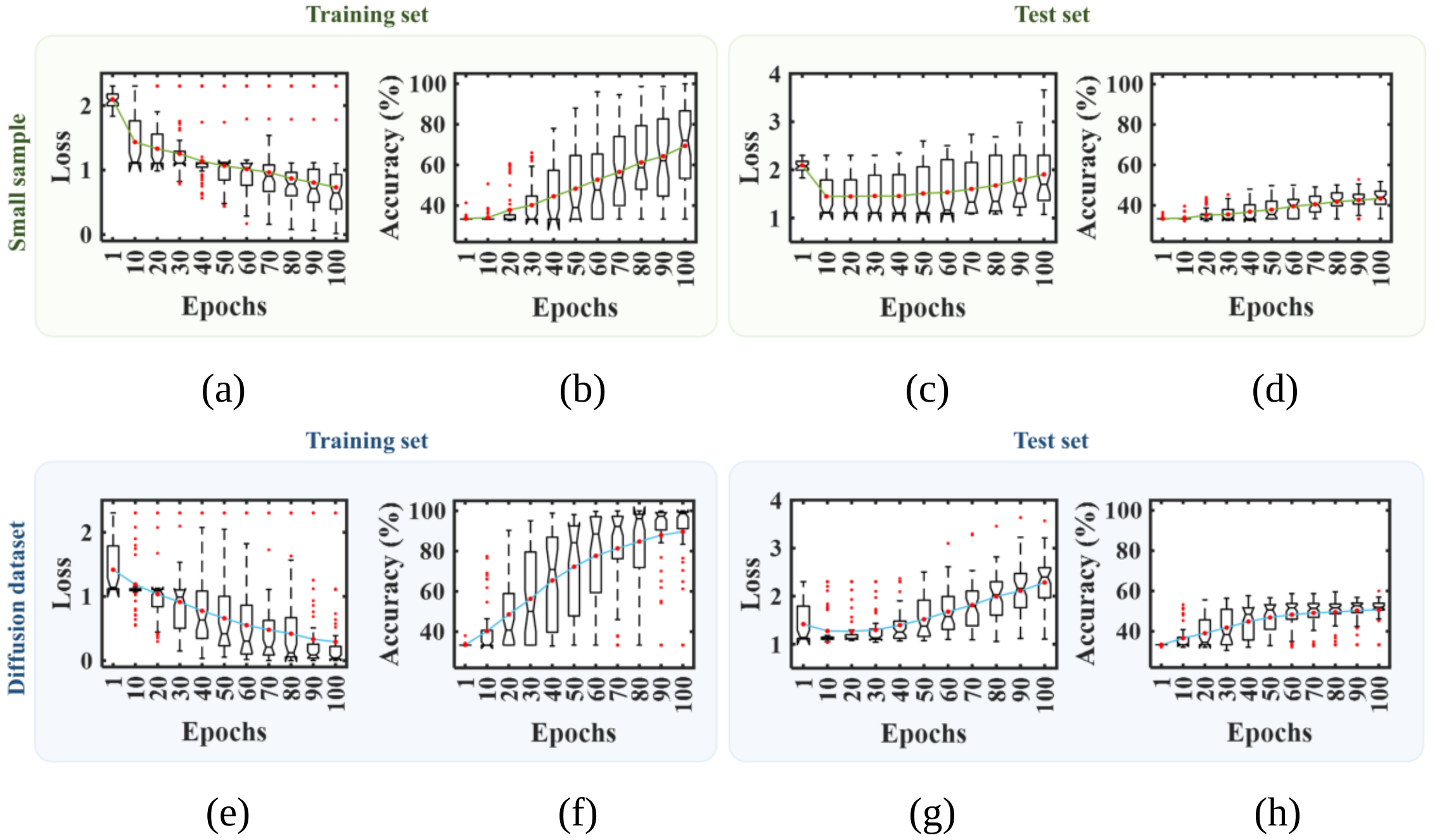

Fig. 34 Box plot of RNNLSTM training process under HIT dataset. (a) Loss function of training set under small sample dataset. (b) Accuracy of training set under small sample dataset. (c) Loss function of test set under small sample dataset. (d) Accuracy of test set under small sample dataset. (e) Loss function of training set under diffusion training set. (f) Accuracy of training set under diffusion training set. (g) Loss function of test set under diffusion training set. (h) Accuracy of test set under diffusion training set.

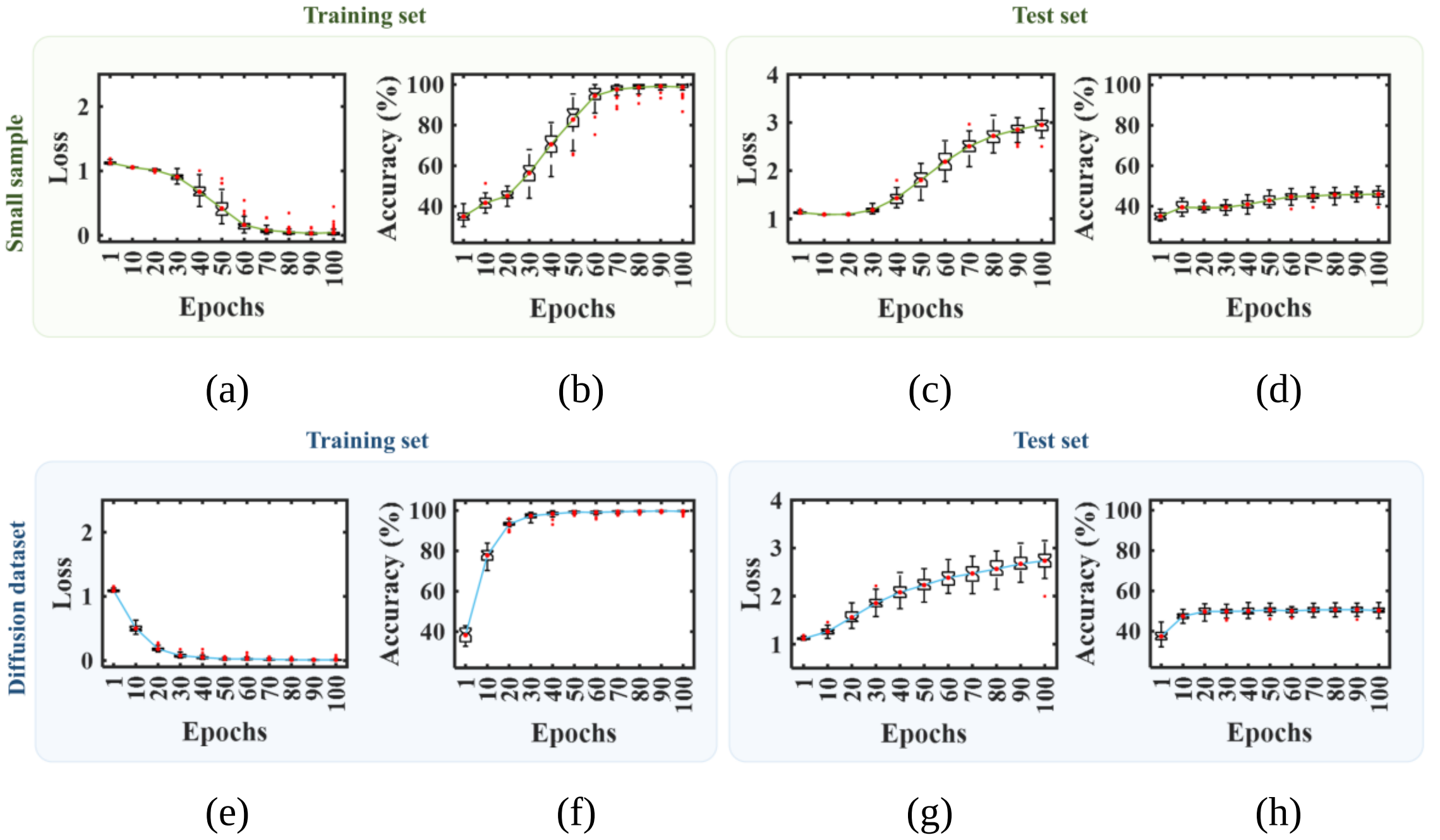

Fig. 35 Box plot of TST training process under HIT dataset. (a) Loss function of training set under small sample dataset. (b) Accuracy of training set under small sample dataset. (c) Loss function of test set under small sample dataset. (d) Accuracy of test set under small sample dataset. (e) Loss function of training set under diffusion training set. (f) Accuracy of training set under diffusion training set. (g) Loss function of test set under diffusion training set. (h) Accuracy of test set under diffusion training set.

From the training process of CNN in Fig. 33, it can be seen that CNN has a good training effect on the loss function and accuracy of HIT small samples and diffusion training set. In the loss function of the test set shown in Fig. 33(c), the loss function increased significantly, which is caused by overfitting. The use of the diffusion training set aggravates the overfitting phenomenon, but it improves the accuracy of the test set.

The training process of RNNLSTM in Fig. 30 shows that the training results based on small sample dataset are seriously discrete, which is reflected in the box plot that the box is too long, especially in Fig. 34(b) and (c). At the same time, the small sample training set causes the loss function and accuracy of the training set not to converge before epoch=100. These problems have been improved after using the diffusion training set. As can be seen from Fig. 34(e) and (f), the loss function and accuracy of the training set are still discrete, which is weaker than before, and the loss function decreases and the accuracy increases. The diffusion training set slightly aggravates the overfitting phenomenon of the loss function of the test set but improves the accuracy.

From the training process of TST in Fig. 35, it can be seen that the results of TST mainly have the problems of slow convergence of loss function and accuracy of the training set, overfitting of the test set and low accuracy of the test set. After using the diffusion training set, the loss function and accuracy of the training set converge faster. From the loss function of the test set in Fig. 35(g), it can be seen that the overfitting phenomenon still exists, but as shown in Fig. 35(h), the accuracy of the test set has been improved.

From the fault diagnosis results of the small sample dataset based on HIT dataset in Fig. 33 to Fig. 35, it can be seen that the overfitting phenomenon is serious. This is because the speed range of HIT dataset is too large, from 1000r/min to 6000r/min. And there are up to 28 speed combinations for each work condition, and the data are measured by six sensors at each speed, including two displacement sensors and four acceleration sensors. Therefore, for the small sample training set dataset with 50 samples of each work condition, there will be a lot of data at the speed and sensor data missing, which will have a much lower probability on the test set with 1200 samples. For the diffusion training set of 1200 samples generated based on the diffusion of 50 samples, it is also difficult to fit enough samples under the speeds and sensors. For the CWRU and XJTU datasets, their speed range is very small, so in most of the machine learning methods used in this paper, TSDM can eliminate the overfitting phenomenon caused by the small number of training set samples. For the above reasons, the overfitting phenomenon of fault diagnosis results of small sample dataset based on HIT dataset is serious, and TSDM cannot effectively eliminate it. But TSDM still improves the accuracy of the test set, which also shows the powerful generation ability of TSDM and its effective supplement to small sample datasets.

The accuracy of the test set at epoch=100 is summarized to reflect the contribution of TSDM and other methods, the box plot is shown in Fig. 36, and the summary table is shown in Tab. 10. It can be seen from Fig. 36 that the application of TSDM to expand the training set can effectively improve the accuracy of small sample fault diagnosis. The other three methods can also improve the accuracy, but the effect is obviously not as good as TSDM. The improvement of proposed TSDM ranges are 9.298%, 7.360 and 4.345% over small sample dataset respectively. The specific diagnostic accuracy results are shown in Tab. 10.

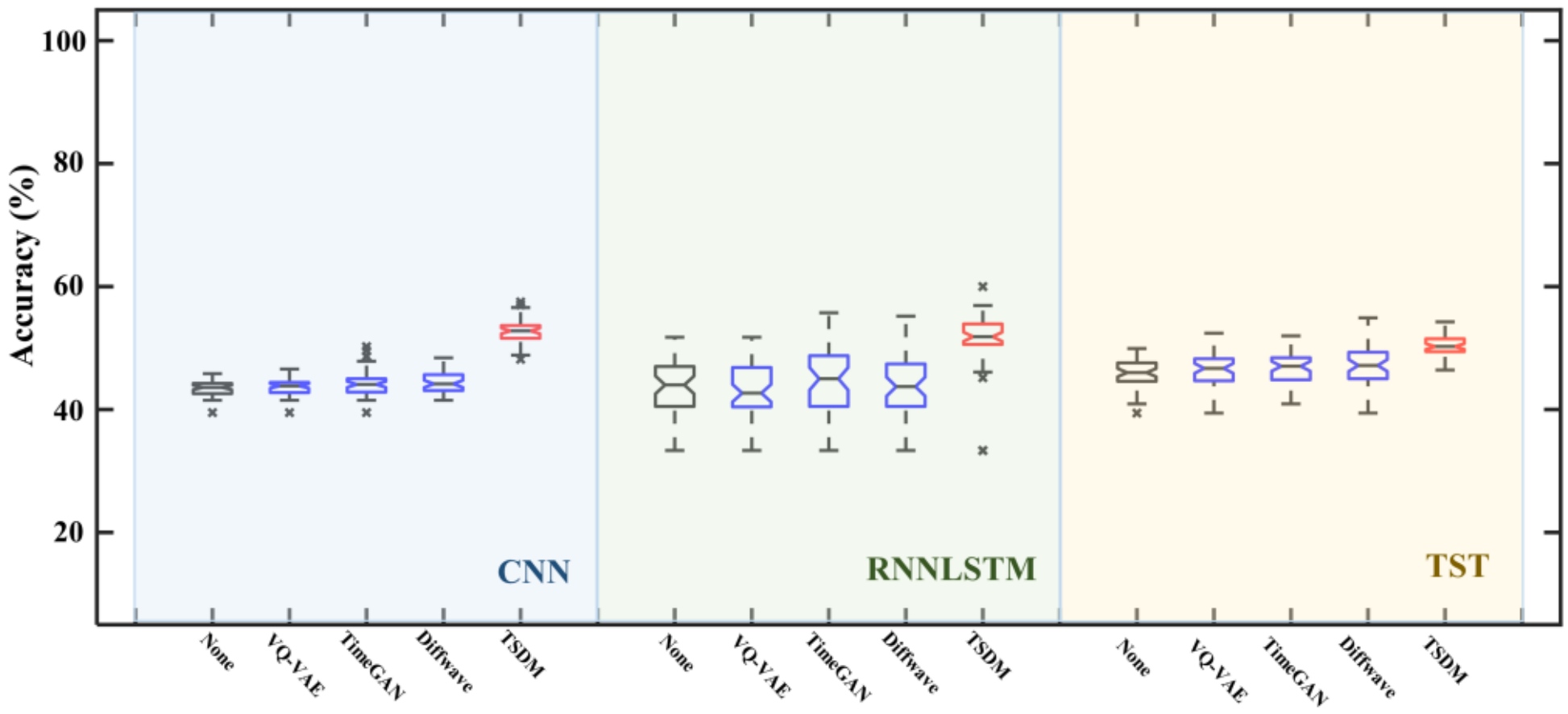

Fig. 36 Box plot of TSDM effect on test set accuracy under HIT dataset at epoch=100.

Tab. 10 The accuracy of test set improved by TSDM under HIT dataset at epoch=100.

| Method | Accuracy of test set | | | | | |
|---|---|---|---|---|---|---|
| | None | VQ-VAE | TimeGAN | Diffwave | TSDM | TSDM improved |
| CNN | 43.455% | 43.833% | 44.083% | 44.167% | 52.753% | 9.298% |
| RNNLSTM | 43.362% | 42.677% | 45.000% | 43.725% | 50.722% | 7.360% |
| TST | 45.945% | 46.696% | 47.042% | 47.167% | 50.290% | 4.345% |

## 6. Conclusions

This paper has proposed a Time Series Denoising Method (TSDM) for time series

generation based on the denoising diffusion probabilistic models. In TSDM, the U-net is improved to make it suitable for the segmentation and feature extraction of one-dimensional time series and is applied to the noise prediction of TSDM. The effectiveness of TSDM is tested on single-frequency, multi-frequency and bearing fault datasets, and TSDM is applied to small sample fault diagnosis. The conclusions are summarized as follows:

(1) TSDM is used to generate single-frequency and multi-frequency artificially constructed trigonometric function datasets. The results show that the periodicity of the generated trigonometric function series is consistent with the original series, and the generated series of the multi-frequency dataset exists in the beat phenomenon similar to the original series. It can be seen from the generated frequency spectrum that the generated time series retains the frequency characteristics of the original series well. Compared with other time series generation methods, TSDM performs better in the frequency accuracy of the generated results.

(2) A public bearing fault dataset is diffused and generated by TSDM. After comparing the frequency spectrums of the generated series with the average spectrums of the original series, the results show that the generated time series frequency spectrums are highly fitted with the average frequency spectrums of the original series, which proves that TSDM can retain the frequency characteristics of the actual vibration signal while generating by diffusion. It also means that TSDM can be applied to fault diagnosis.

(3) Based on three public bearing fault diagnosis datasets, CWRU, XJTU and HIT datasets, a case of small sample fault diagnosis is defined. And TSDM is used to generate the small sample training set to expand the dataset. The results show that when using CNN, LSTM and TST for small sample fault diagnosis of the three datasets, the diffusion datasets generated by TSDM can effectively improve the accuracy of small sample fault diagnosis, with a maximum increase of 32.380%. Compared with other time series generation methods, TSDM performs better in the accuracy of fault diagnosis.

The results of this paper show that the proposed TSDM model has a solid ability to generate time series, and in terms of generating vibration signals, it is significantly better than other generation methods. Future work will focus on optimizing the TSDM model and improving its fault diagnosis accuracy for small sample datasets. The small sample fault diagnosis using TSDM in this paper is not comprehensive enough, the model can be further improved to improve the accuracy of small sample fault diagnosis.

## Declaration of competing interest

The authors declare that they have no known competing financial interests or personal relationships that could have appeared to influence the work reported in this paper.

## Acknowledgment

The authors are very grateful for the financial supports from the National Key R&D Program of China (Grant No. 2023YFE0125900), National Natural Science Foundation of China (Grant No. 12372008), Natural Science Foundation of Heilongjiang Province (Grant No. YQ2022A008) and the Fundamental Research Funds for the Central Universities, Polish National Science Centre, Poland under the grant OPUS 18 No. 2019/35/B/ST8/00980.

## References

[1] S. Nandi, H. Toliyat, X. Li, Condition monitoring and fault diagnosis of electrical motorsa review, IEEE Transactions on Energy Conversion 20 (2005) 719–729.

[2] X. Jiang, C. Shen, J. Shi, Z. Zhu, Initial center frequency-guided vmd for fault diagnosis of rotating machines, Journal of Sound and Vibration 435 (2018) 36–55

[3] A. Glowacz, R. Tadeusiewicz, S. Legutko, W. Caesarendra, M. Irfan, H. Liu, F. Brumercik, M. Gutten, M. Sulowicz, J. A. Antonino Daviu, T. Sarkodie-Gyan, P. Fracz, A. Kumar, J. Xiang, Fault diagnosis of angle grinders and electric impact drills using acoustic signals, Applied Acoustics 179 (2021) 108070.

[4] X. Yan, M. Jia, A novel optimized svm classification algorithm with multi-domain feature and its application to fault diagnosis of rolling bearing, Neurocomputing 313 (2018) 47–64.

[5] N. Saeed, E. Awwad, M. El-Meligy, E. Nasr. Radial versus Cartesian control strategies to stabilize the nonlinear whirling motion of the six-pole rotor-AMBs, IEEE Access 8 (2020) 138859-138883.

[6] N. Saeed, E. Mahrous, J. Awrejcewicz. Nonlinear dynamics of the six-pole rotor-AMB system under two different control configurations, Nonlinear Dynamics 101 (4), (2020) 2299-2323.

[7] D. Xiao, J. Ding, X. Li, L. Huang, Gear fault diagnosis based on kurtosis criterion vmd and som neural network, Applied Sciences 9 (2019) 5424–5449.

[8] W. Mao, W. Feng, Y. Liu, D. Zhang, X. Liang, A new deep auto-encoder method with fusing discriminant information for bearing fault diagnosis, Mechanical Systems and Signal Processing 150 (2021) 107233.

[9] S. Kiranyaz, O. Avci, O. Abdeljaber, T. Ince, M. Gabbouj, D. J. Inman, 1d convolutional neural networks and applications: A survey, Mechanical Systems and Signal Processing 151 (2021) 107398.

[10] R. Bai, Q. Xu, Z. Meng, L. Cao, K. Xing, F. Fan, Rolling bearing fault diagnosis based on multi-channel convolution neural network and multi-scale clipping fusion data augmentation,

Measurement 184 (2021) 109885.
[11] Y. Zhang, X. Li, L. Gao, L. Wang, L. Wen, Imbalanced data fault diagnosis of rotating machinery using synthetic oversampling and feature learning, Journal of Manufacturing Systems 48 (2018) 34–50. Special Issue on Smart Manufacturing.
[12] T. Zheng, L. Song, J. Wang, W. Teng, X. Xu, C. Ma, Data synthesis using dual discriminator conditional generative adversarial networks for imbalanced fault diagnosis of rolling bearings, Measurement 158 (2020) 107741.
[13] W. Wan, S. He, J. Chen, A. Li, Y. Feng, Qscgan: An un-supervised quick self-attention convolutional gan for lre bearing fault diagnosis under limited label-lacked data, IEEE Transactions on Instrumentation and Measurement 70 (2021) 1–16.
[14] H. Wang, Z. Liu, T. Ai, Long-range dependencies learning based on non-local 1d-convolutional neural network for rolling bearing fault diagnosis, Journal of Dynamics, Monitoring and Diagnostics (2022).
[15] DP. Kingma and M. Welling. Auto-encoding variational Bayes. ArXiv:1312.6114, (2013).
[16] T. Salimans, DP. Kingma and M. Welling. Markov chain Monte Carlo and variational inference: bridging the gap. In International Conference on Machine Learning, Beijing, China, 21–26 June (2014), 1218–1226.
[17] DJ. Rezende, S. Mohamed and D. Wierstra. Stochastic backpropagation and approximate inference in deep generative models, ArXiv:1401.4082, (2014).
[18] G. Turinici. Radon Sobolev Variational Auto-Encoders. DOI:10.48550/arXiv.1911.13135. (2019).
[19] N. Anand and T. Achim. Protein Structure and Sequence Generation with Equivariant Denoising Diffusion Probabilistic Models. arXiv preprint arXiv:2205.15019 (2022).
[20] E. Asiedu, S. Kornblith, T. Chen, N. Parmar, M. Minderer, and M. Norouzi. Decoder Denoising Pretraining for Semantic Segmentation. ArXiv abs/2205.11423 (2022).
[21] J. Sohl-Dickstein, E. Weiss, N. Maheswaranathan, S. Ganguli, Deep unsupervised learning using nonequilibrium thermodynamics. In International Conference on Machine Learning, (2015) 2256-2265.
[22] J. Ho, A. Jain, P. Abbeel. Denoising Diffusion Probabilistic Models, DOI: 10.48550/arXiv.2006.11239, (2020).
[23] O. Ronneberger, P. Fischer, T. Brox. U-Net: Convolutional Networks for Biomedical Image Segmentation. Springer International Publishing, (2015). DOI: 10.1007/978-3-319-24574-4_28.
[24] D. Baranchuk, I. Rubachev, A. Voynov, V. Khrulkov, and A. Babenko. Label-Efficient Semantic Segmentation with Diffusion Models. ArXiv abs/2112.03126 (2022).
[25] C. Saharia, W. Chan, H. Chang, C. Lee, J. Ho, T. Salimans, D. Fleet, and M. Norouzi.. Palette: Image-to-image diffusion models. In Special Interest Group on Computer Graphics and Interactive Techniques Conference Proceedings. (2022) 1–10.
[26] G. Batzolis, J. Stanczuk, C. Schönlieb, and C. Etmann. Conditional image generation with score-based diffusion models. arXiv preprint arXiv:2111.13606 (2021)
[27] R. Yang, Y. Yang, J. Marino, and S. Mandt. 2021. Insights from Generative Modeling for Neural Video Compression. arXiv preprint arXiv:2107.13136 (2021)
[28] R. Rombach, A. Blattmann, D. Lorenz, P. Esser, and B. Ommer. High-Resolution Image Synthesis with Latent Diffusion Models. ArXiv abs/2112.10752 (2021)
[29] R. Yang, P. Srivastava, and S. Mandt. Diffusion probabilistic modeling for video generation. arXiv preprint arXiv:2203.09481 (2022).

[30] T. Chen, R. Zhang, and G. Hinton. Analog Bits: Generating Discrete Data using Diffusion Models with Self-Conditioning. arXiv preprint arXiv:2208.04202 (2022).
[31] J. Austin, D. Johnson, J. Ho, D. Tarlow, and R. Berg. Structured denoising diffusion models in discrete state-spaces. Advances in Neural Information Processing Systems 34 (2021), 17981–17993.
[32] T. Chen, R. Zhang, and G. Hinton. Analog Bits: Generating Discrete Data using Diffusion Models with Self-Conditioning. arXiv preprint arXiv:2208.04202 (2022).
[33] N. Chen, Y. Zhang, H. Zen, R. Weiss, Mohammad Norouzi, and William Chan. WaveGrad: Estimating Gradients for Waveform Generation. ArXiv abs/2009.00713 (2021).
[34] Z. Kong, W. Ping, J. Huang, K. Zhao, and B. Catanzaro. DiffWave: A Versatile Diffusion Model for Audio Synthesis. ArXiv abs/2009.09761 (2021).
[35] B. Jing, G. Corso, R. Barzilay, and T. Jaakkola. Torsional Diffusion for Molecular Conformer Generation. In ICLR2022 Machine Learning for Drug Discovery. (2022).
[36] M. Xu, L. Yu, Y. Song, Ch. Shi, S. Ermon, and J. Tang. GeoDiff: A Geometric Diffusion Model for Molecular Conformation Generation. In International Conference on Learning Representations. (2021).
[37] B. Trippe, J. Yim, D. Tischer, T. Broderick, D. Baker, R. Barzilay, and T. Jaakkola. Diffusion probabilistic modeling of protein backbones in 3D for the motif-scaffolding problem. arXiv preprint arXiv:2206.04119 (2022).
[38] C. Shi, S. Luo, M. Xu, and J. Tang. Learning gradient fields for molecular conformation generation. In International Conference on Machine Learning. (2021), PMLR, 9558–9568.
[39] J. Yoon, S. Huang, and J. Lee. Adversarial purification with score-based generative models. In International Conference on Machine Learning. (2021). PMLR, 12062–12072.
[40] T. Blau, R. Ganz, B. Kawar, A. Bronstein, and M. Elad. Threat Model-Agnostic Adversarial Defense using Diffusion Models. arXiv abs/2207.08089 (2022).
[41] Q. Wu, H. Ye, and Y. Gu. Guided Diffusion Model for Adversarial Purification from Random Noise. arXiv preprint arXiv:2206.10875 (2022).
[42] A. Oord, O. Vinyals, K. Kavukcuoglu. Neural Discrete Representation Learning (2017). arXiv preprint arXiv: 1711.00937.
[43] Yoon, Jinsung, D. Jarrett, and M. Schaar. Time-series generative adversarial networks[J]. Advances in neural information processing systems, 2019, 32.
[44] Z. Kong, W. Ping, J. Huang, K. Zhao, B. Catanzaro. DiffWave: A Versatile Diffusion Model for Audio Synthesis (2020). arXiv preprint arXiv: 2009.09761.
[45] X. Li, W. Zhang, Q. Ding, J. Sun, Intelligent rotating machinery fault diagnosis based on deep learning using data augmentation, Journal of Intelligent Manufacturing 31 (2020) 433–452.
[46] L. Ma, Y. Ding, Z. Wang, C. Wang, J. Ma, C. Lu, An interpretable data augmentation scheme for machine fault diagnosis based on a sparsity-constrained generative adversarial network, Expert Systems with Applications 182 (2021) 115234.
[47] M. Talab, S. Awang, S. Najim, Super-low resolution face recognition using integrated efficient sub-pixel convolutional neural network (espcn) and convolutional neural network (cnn), in: 2019 IEEE International Conference on Automatic Control and Intelligent Systems (I2CACIS), (2019), 331–335.
[48] X. Liu, G. Chen, T. Hao. A combined deep learning model for damage size estimation of rolling bearing, International J of Engine Research, 24 (2023), 1362–1373.

[49] W. Jung, S. Kim, S. Yun. Vibration, acoustic, temperature, and motor current dataset of rotating machine under varying operating conditions for fault diagnosis, Data in Brief, 48 (2023) Art.ID.109049.

[50] Y. Jin, L. Hou, Y. Chen. A Time Series Transformer based method for the rotating machinery fault diagnosis, Neurocomputing, 494, (2022) 379–395.

[51] A. Smith, B. Randall. Rolling element bearing diagnostics using the Case Western Reserve University data: A benchmark study, Mechanical Systems and Signal Processing, 64-65, (2015) 100-131.

[52] B. Wang, Y. Lei, N. Li. A hybrid prognostics approach for estimating remaining useful life of rolling element bearings, IEEE Transactions on Reliability, 69, (2020) 401-412.

[53] L. Hou, H. Yi, Y. Jin, M. Gui, L. Sui, J. Zhang, Y. Chen. Inter-shaft Bearing Fault Diagnosis Based on Aero-engine System: A Benchmarking Dataset Study. Journal of Dynamics, Monitoring and Diagnostics. (2023). https://doi.org/10.37965/jdmd.2023.314.

[54] H. Yi, L. Hou, P. Gao. Nonlinear resonance characteristics of a dual-rotor system with a local defect on the inner ring of the inter-shaft bearing, Chinese Journal of Aeronautics, 34, (2021) 110-124.

[55] M. Nassar, N. Saeed, A. Nasedkin. Determination of effective properties of porous piezoelectric composite with partially randomly metalized pore boundaries using finite element method. Applied Mathematical Modelling 124 (2023) 241-256.